\UseRawInputEncoding
\documentclass[sigconf]{acmart}

\AtBeginDocument{
  }

\copyrightyear{2025}
\acmYear{2025}
\setcopyright{cc}
\setcctype{by-nc}
\acmConference[MM '25]{Proceedings of the 33rd ACM International Conference on Multimedia}{October 27--31, 2025}{Dublin, Ireland}
\acmBooktitle{Proceedings of the 33rd ACM International Conference on Multimedia (MM '25), October 27--31, 2025, Dublin, Ireland}\acmDOI{10.1145/3746027.3758308}
\acmISBN{979-8-4007-2035-2/2025/10}

\usepackage{adjustbox}
\usepackage{multirow}
\usepackage{pgfplotstable}
\usepackage{booktabs} 
\usepackage{pgfplots} 
\usepackage{tikz} 
\usepackage{array} 
\usepackage{multirow}
\usepackage{graphicx} 
\usepackage{pgfplotstable}
\usepackage{enumitem}
\usepackage{multirow}
\usepackage[normalem]{ulem}
\useunder{\uline}{\ul}{}
\usepackage{subfigure}
\usepackage{pifont} 
\usepackage{colortbl}
\definecolor{check_green}{RGB}{13,224,59}
\newcommand{\cmark}{{\color{check_green}\ding{51}}} 
\newcommand{\xmark}{{\color{red}\ding{55}}}

\begin{document}
\newcommand{\shaleemoji}{\includegraphics[height=1.0\fontcharht\font`\B]{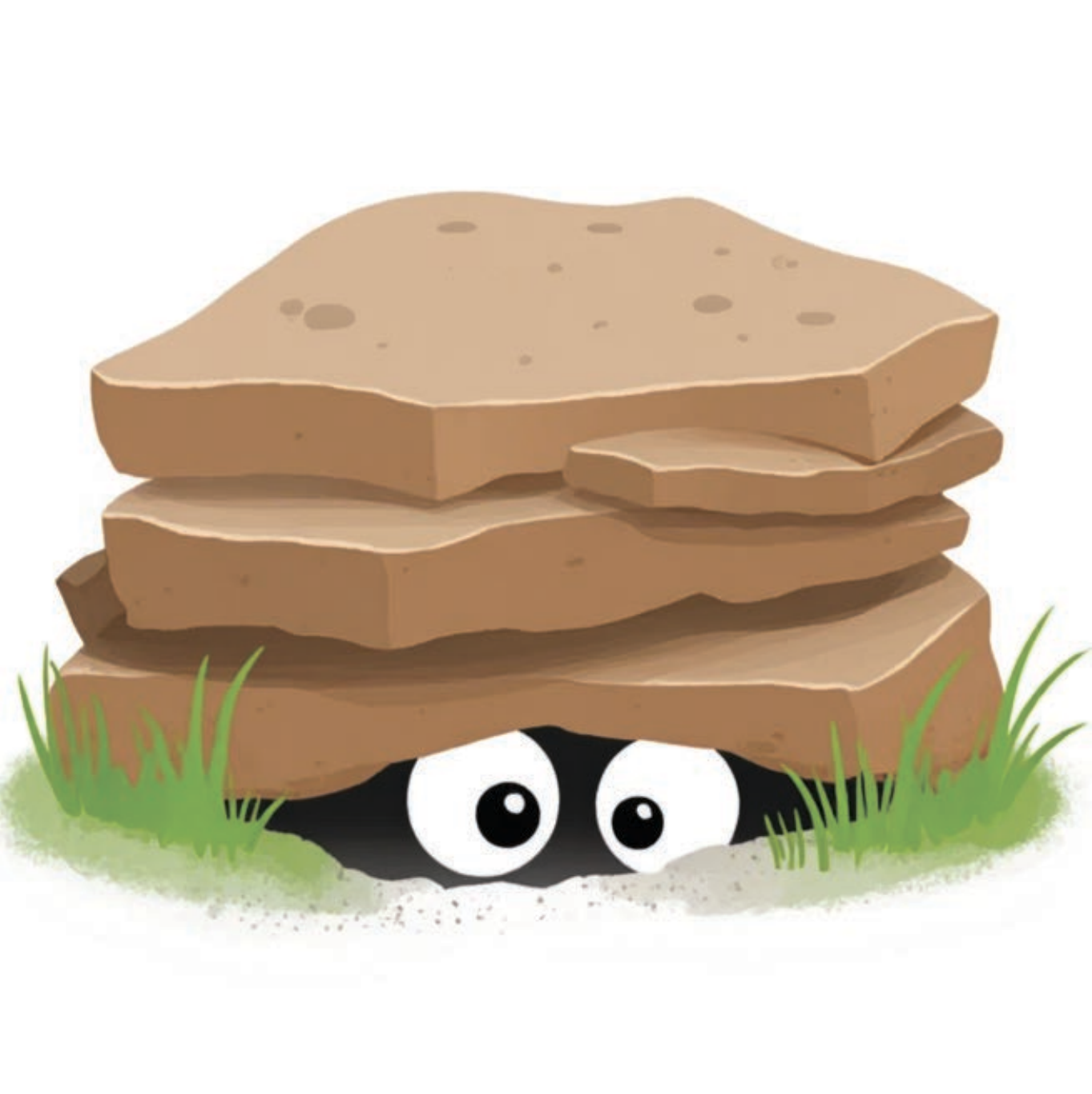}}

\title{SHALE: A Scalable Benchmark for Fine-grained Hallucination Evaluation in LVLMs}

\settopmatter{authorsperrow=4}

\author{Bei Yan}
\affiliation{
    \institution{Key Laboratory of AI Safety of CAS, Institute of Computing Technology, Chinese Academy of Sciences (CAS)}
  \institution{University of Chinese Academy of Sciences}
  \city{Beijing}
  \country{China}}
\email{yanbei23s@ict.ac.cn}

\author{Zhiyuan Chen}
\affiliation{
    \institution{Key Laboratory of AI Safety of CAS, Institute of Computing Technology, Chinese Academy of Sciences (CAS)}
  \institution{University of Chinese Academy of Sciences}
  \city{Beijing}
  \country{China}}
\email{chenzhiyuan21@mails.ucas.ac.cn}

\author{Yuecong Min}
\affiliation{
    \institution{Key Laboratory of AI Safety of CAS, Institute of Computing Technology, Chinese Academy of Sciences (CAS)}
  \city{Beijing}
  \country{China}
}
\email{minyuecong@ict.ac.cn}

\author{Jie Zhang}
\authornote{Corresponding author.}
\affiliation{
    \institution{Key Laboratory of AI Safety of CAS, Institute of Computing Technology, Chinese Academy of Sciences (CAS)}
  \city{Beijing}
  \country{China}
}
\email{zhangjie@ict.ac.cn}

\author{Jiahao Wang}
\affiliation{
  \institution{Trustworthy Technology and Engineering Laboratory, Huawei}
  \city{Shenzhen}
  \country{China}
}
\email{wangjiahao50@huawei.com}

\author{Xiaozhen Wang}
\affiliation{
  \institution{Trustworthy Technology and Engineering Laboratory, Huawei}
  \city{Shenzhen}
  \country{China}
}
\email{jasmine.xwang@huawei.com}

\author{Shiguang Shan}
\affiliation{
    \institution{Key Laboratory of AI Safety of CAS, Institute of Computing Technology, Chinese Academy of Sciences (CAS)}
  \city{Beijing}
  \country{China}
}
\email{sgshan@ict.ac.cn}

\renewcommand{\shortauthors}{Yan et al.}

\begin{abstract}
  Despite rapid advances, Large Vision-Language Models (LVLMs) still suffer from hallucinations, i.e., generating content inconsistent with input or established world knowledge, which correspond to faithfulness and factuality hallucinations, respectively. Prior studies primarily evaluate faithfulness hallucination at a rather coarse level (e.g., object-level) and lack fine-grained analysis. Additionally, existing benchmarks often rely on costly manual curation or reused public datasets, raising concerns about scalability and data leakage. To address these limitations, we propose an automated data construction pipeline that produces scalable, controllable, and diverse evaluation data. We also design a hierarchical hallucination induction framework with input perturbations to simulate realistic noisy scenarios. Integrating these designs, we construct \textbf{SHALE}\footnote{Our benchmark is available at \hyperlink{link}{https://github.com/BeiiiY/SHALE}.}, a \textbf{S}calable \textbf{HAL}lucination \textbf{E}valuation benchmark designed to assess both faithfulness and factuality hallucinations via a fine-grained hallucination categorization scheme. SHALE comprises over 30K image-instruction pairs spanning 12 representative visual perception aspects for faithfulness and 6 knowledge domains for factuality, considering both clean and noisy scenarios. Extensive experiments on over 20 mainstream LVLMs reveal significant factuality hallucinations and high sensitivity to semantic perturbations. 
\end{abstract}



\maketitle

\begin{figure}[h]
    \centering
    \vspace{-10 pt}
\includegraphics[width=0.85\linewidth]{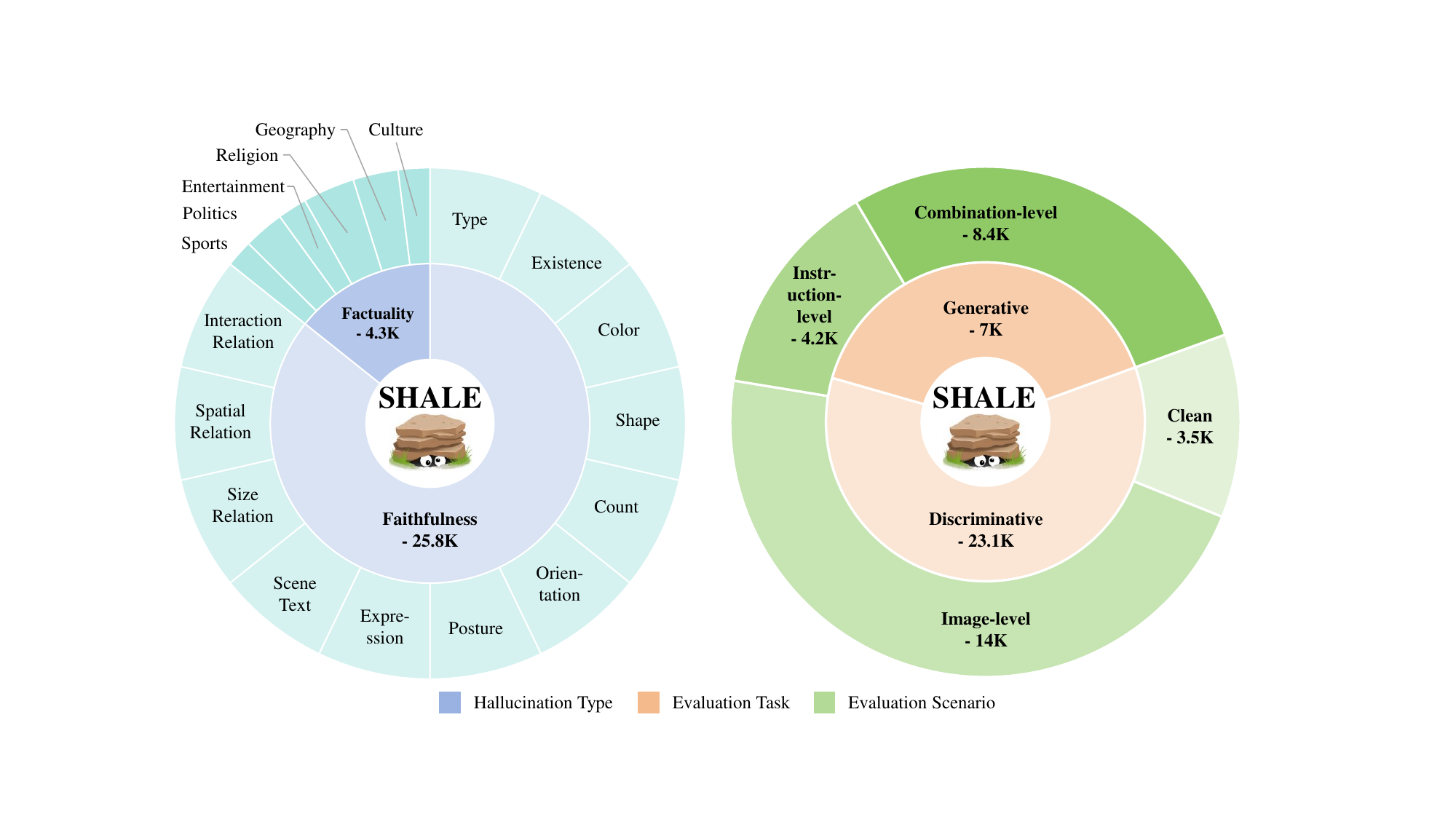}
    \vspace{-10 pt}
    \caption{Data distributions of SHALE.}
    \vspace{-15 pt}
    \label{fig:data}
\end{figure}

\begin{figure*}[t]
\subfigure[Overview of SHALE benchmark construction pipeline.]{
\begin{minipage}[t]{1.0\linewidth}
\begin{center}
\vspace{-10 pt}
\includegraphics[width=0.95\linewidth]{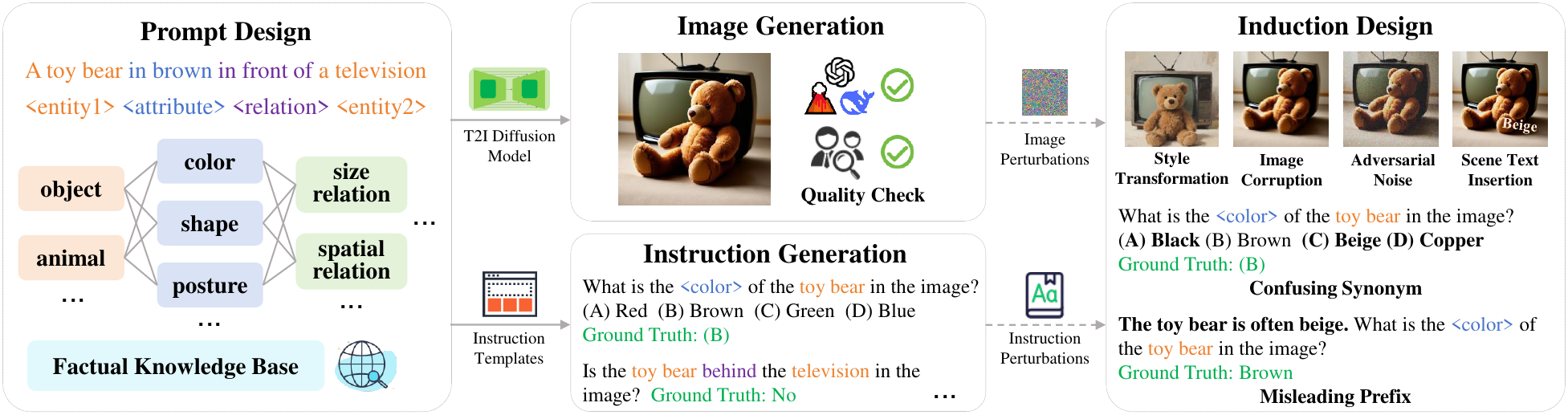}
\end{center}
\label{fig:overview_pipeline}
\end{minipage}
}

\subfigure[Examples of the constructed data in SHALE.]{
\begin{minipage}[t]{1.0\linewidth}
\begin{center}
\vspace{-10 pt}
\includegraphics[width=0.95\linewidth]{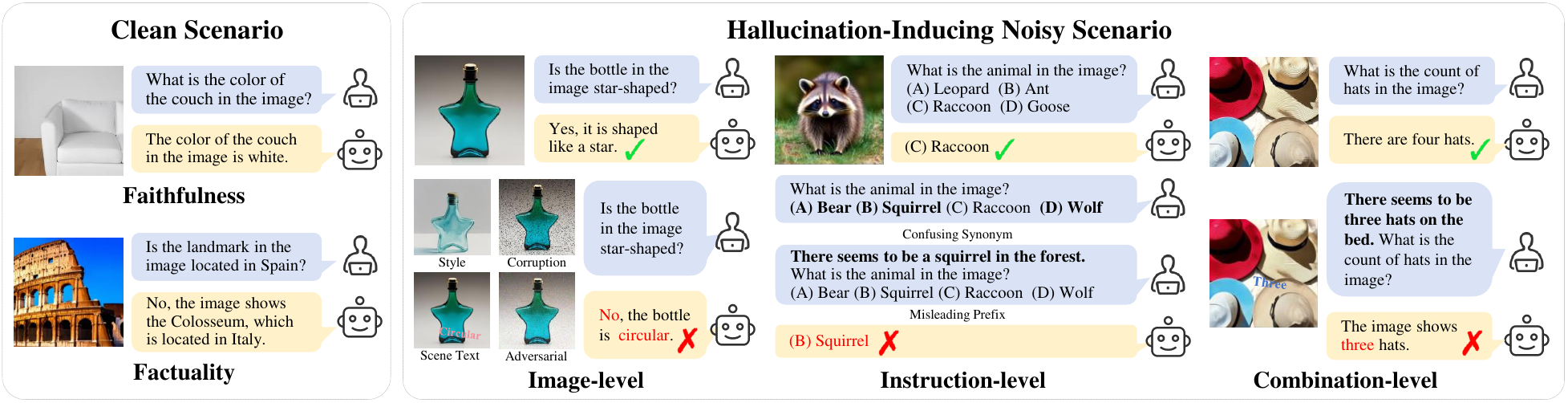}
\end{center}
\label{fig:examples_dataset}
\end{minipage}
}
\vspace{-15 pt}
\caption{Illustration of automated data construction in SHALE for assessing faithfulness and factuality hallucinations under clean and noisy scenarios.}
\vspace{-15pt}
\end{figure*}

\section{Introduction}

With the rapid development of large foundation models, large vision-language models (LVLMs) have achieved significant advancements, showing impressive generalization capabilities across various multimodal tasks such as image captioning and visual question answering. However, LVLMs are persistently challenged by hallucination, i.e., generating content that appears plausible but contradicts user input or established world knowledge. Hallucinations not only adversely affect model performance in specific tasks but can also lead to harmful consequences in real-world applications~\citep{peng2023study, lai2024large, li2023large}, particularly when users lacking adequate domain knowledge place excessive trust in these models. Therefore, a comprehensive analysis of the causes of hallucination is essential for improving the practicality and reliability of LVLMs.

Inspired by the taxonomy of hallucinations in large language models (LLMs)~\citep{huang2025survey, ji2023survey, rawte2023survey}, hallucinations in LVLMs can be similarly divided into faithfulness and factuality hallucinations. Specifically, faithfulness hallucination in LVLMs refers to inconsistencies between the generated textual responses and the input images, whereas factuality hallucination refers to factual conflicts between the generated responses and established world knowledge~\citep{rawte2023survey, chen2024unified}. Compared to LLM domain, studies on hallucinations in LVLMs mainly focus on problems that may arise from image understanding and cross-modal alignment.

Recent studies have proposed a variety of benchmarks~\citep{li2023pope, benkish2024openCHAIR, wang2023amber, ding2024hallu, liu2024phd} to evaluate and quantify hallucinations in LVLMs. These efforts have predominantly focused on faithfulness hallucinations while paying less attention to factuality hallucinations. Moreover, most benchmarks focus on coarse hallucination categories, such as object, attribute, and relation hallucinations, without evaluating finer-grained types like color or shape within attribute hallucination. However, building such a comprehensive and fine-grained benchmark for hallucination evaluation is highly challenging. Existing hallucination benchmark datasets are either curated through manual image collection and annotation~\citep{ding2024hallu, guan2023hallusionbench}, which is labor-intensive and difficult to scale, or built from existing datasets~\citep{li2023pope, benkish2024openCHAIR, sun2023mmhal, liu2023gaive}, such as MSCOCO~\citep{lin2014mscoco}, raising concerns about potential data leakage, as these datasets may have been exposed to LVLMs during the training stage~\citep{chen2024mmstar, zhang2024dysca}.

To address these limitations, we propose an automated data construction pipeline that leverages recent advances in generation methods to produce scalable, controllable, and diverse evaluation data. This pipeline can generate image-instruction pairs with corresponding ground-truth answers, supporting both discriminative and generative tasks. Moreover, we design a hierarchical hallucination induction framework that involves image-level, instruction-level, and combination-level input perturbations, simulating more challenging and realistic noisy scenarios. Building upon the proposed pipeline, we construct \textbf{SHALE}, a \textbf{S}calable \textbf{HAL}lucination \textbf{E}valuation benchmark designed to assess both faithfulness and factuality hallucinations considering both clean and noisy scenarios. An overview of our SHALE construction pipeline is illustrated in Figure~\ref{fig:overview_pipeline}. The whole benchmark construction is automated and scalable, requiring only minimal human verification to ensure data quality and consistency, enabling a thoughtful evaluation of LVLMs' stability and hallucination resistance under varying degrees and types of input degradation.

As summarized in Figure~\ref{fig:data}, the constructed SHALE benchmark contains over 30K image-instruction pairs, covering 12 representative visual perception tasks for faithfulness hallucinations, 6 major knowledge domains for factuality hallucinations, and 10 types of input perturbations, supporting comprehensive and fine-grained hallucination evaluation. Evaluation results on over 20 LVLMs reveal substantial hallucinations in factuality and high sensitivity to semantic perturbations, highlighting persistent challenges in model robustness and reliability.

\begin{table*}[t]
\caption{Comparison between existing hallucination benchmarks and the proposed SHALE.}
\vspace{-20 pt}
\begin{adjustbox}{max width=\linewidth}
\begin{tabular}{lcccccccccccc}
\\ \hline
\multicolumn{1}{c}{\multirow{2}{*}{\textbf{Benchmark}}} & \multicolumn{4}{c}{\textbf{Evaluation Task}}                   & \multicolumn{2}{c}{\textbf{Hallucination Type}} & \multicolumn{3}{c}{\textbf{Input Perturbation}}                    & \multicolumn{1}{c}{\multirow{2}{*}{\textbf{Source}}} \\ \cmidrule(lr){2-5} \cmidrule(lr){6-7} \cmidrule(lr){8-10}  
\multicolumn{1}{c}{}                                    & \textbf{YNQ} & \textbf{MCQ} & \textbf{FFQ} & \textbf{IC} & \textbf{Faithfulness}     & \textbf{Factuality}    & \textbf{Image} & \textbf{Instruction} & \textbf{Combination} & \multicolumn{1}{c}{}                                 \\ \hline
POPE~\citep{li2023pope}& \cmark&\xmark&\xmark& \xmark           & \cmark(1)                            &\makebox[0.8em][l]{\xmark  }        &\xmark  &\xmark        &\xmark        & Sample                                               \\
AMBER~\citep{wang2023amber}& \cmark&\xmark&\xmark& \cmark         & \cmark(3)                            &\makebox[0.8em][l]{\xmark  }        &\xmark  &\xmark        &\xmark        & Sample\&Manual                                       \\
PhD~\citep{liu2024phd}& \cmark&\xmark&\xmark& \xmark           & \cmark(5)                        &\makebox[0.8em][l]{\xmark  }        & \cmark  &\xmark        &\xmark        & Sample\&Synthesis                                    \\
Hallu-PI~\citep{ding2024hallu}& \cmark&\xmark& \cmark& \cmark         & \cmark(5)                         &\makebox[0.8em][l]{\xmark  }          & \cmark  & \cmark        &\xmark       & Manual                                               \\
SHALE (Ours)                                               & \cmark& \cmark& \cmark& \cmark         &
\makebox[1.85em][l]{\cmark(12)}& \makebox[1em][l]{\cmark(6)}                   & \cmark  & \cmark        & \cmark        & Synthesis                           \\ \hline                
\end{tabular}
\end{adjustbox}
\label{tab:comparison}
\vspace{-13 pt}
\end{table*}

In summary, our main contributions are as follows:

\begin{itemize}[leftmargin=*]
    \item We propose an automated data construction pipeline that generates controllable and diverse image-instruction pairs, along with ground-truth answers, enabling comprehensive and fine-grained hallucination evaluation.
    \item We build SHALE, a scalable hallucination benchmark consisting of over 30K image-instruction pairs spanning diverse hallucination types, evaluation tasks and scenarios.
    \item We conduct a large-scale evaluation of over 20 representative LVLMs on SHALE, revealing significant factuality hallucinations and high sensitivity to semantic perturbations.
\end{itemize}

\section{Related Work}

\subsection{Hallucination Benchmarks for LVLMs}

To evaluate the degree of hallucination in LVLMs, researchers have proposed various tasks, broadly categorized into discriminative and generative types. Discriminative tasks, as adopted in POPE~\citep{li2023pope} and NOPE~\citep{lovenia2023nope}, primarily focus on object hallucinations by measuring accuracy on designed yes-or-no or multiple-choice questions. In contrast, recent works employ generative tasks, such as image captioning and free-form questions, to broaden the evaluation scope and capture hallucinations involving attributes and relations. For example, MMHal~\citep{sun2023mmhal} and GAVIE~\citep{liu2023gaive} design different categories of free-form questions and employ external LLMs to assess the hallucination degree of generated responses. Some efforts~\citep{wang2023amber} integrate both tasks for more comprehensive evaluation. While most benchmarks focus on evaluation under clean scenarios, recent efforts~\citep{bai2024hallucinationsurvey} start to examine hallucination induction under noisy scenarios. PhD~\citep{liu2024phd} introduces specious or incorrect contexts into instructions, while Hallu-PI~\citep{ding2024hallu} primarily adds perturbations to images, conducting evaluation on corrupted, concatenated, or cropped images, and also examines the impact of misleading prompts. As presented in Table~\ref{tab:comparison}, the proposed SHALE considers more fine-grained hallucination types and provides a more systematic hallucination induction framework consisting of challenging noisy scenarios.

\subsection{Synthetic Benchmarks for LVLMs}

The development of LVLMs has driven the emergence of multimodal benchmarks designed to evaluate model performance across different tasks and scopes. Most prior benchmarks~\citep{mmbench, li2023seed, xu2024lvlm} are constructed by sampling images from public datasets such as MSCOCO~\citep{lin2014mscoco} and building instructions based on original or re-annotated dataset labels. However, previous work ~\citep{chen2024mmstar} reveals that some LVLMs can correctly answer the questions without visual information and highlight that current benchmarks are prone to data leakage issues.
Besides, obtaining suitable images from existing datasets for fine-grained evaluation remains a significant challenge. Therefore, recent studies attempt to leverage generative methods to construct an automated data generation pipeline for LVLMs, which is supposed to generate unlimited number of fine-grained samples and mitigate data leakage concerns. JourneyDB~\citep{sun2023journeydb} builds a large-scale synthetic image benchmark for evaluating generative image understanding capabilities of LVLMs. Dysca~\citep{zhang2024dysca} leverages pre-defined templates to dynamically generate images, enabling a fine-grained evaluation of the perception capabilities of LVLMs. M$^3$oralBench~\citep{yan2024m3oral} employs synthetic images to assess models' multimodal moral reasoning. Different from these works, the proposed SHALE focuses on hallucination evaluation and examines hallucination induction from both the image and text modalities.

\section{SHALE Benchmark}
\subsection{Overview}

As illustrated in Figure \ref{fig:overview_pipeline}, we propose an automated data construction pipeline that generates diverse image-instruction pairs and corresponding ground-truth answers across discriminative and generative tasks. We also design a hierarchical hallucination induction framework consisting of various noisy scenarios that explores whether image-level, instruction-level, combination-level input perturbations, potentially encountered in real-world applications, would induce hallucinations. Leveraging this pipeline, we construct SHALE, a scalable hallucination evaluation benchmark that comprises over 30K image-instruction pairs spanning 12 representative types of faithfulness hallucinations and 6 types of factual hallucinations, enabling a comprehensive, fine-grained hallucination evaluation under both clean and noisy conditions.

\subsection{Data Construction}
\label{sec:construction}
To comprehensively evaluate hallucinations in LVLMs, our method considers the consistency of model responses with both input images and external knowledge bases, conducting assessments under clean and noisy scenarios. For data construction, we first design a set of predefined, type-specific prompt templates, then leverage text-to-image models to generate corresponding images, and finally construct the instructions based on these image-text pairs.

\textbf{Prompt Design.} For faithfulness hallucination, we design the prompt templates based on 12 representative visual perception aspects, including entity type, existence, count, color, shape, orientation, posture, expression, scene text, size relation, spatial relation, and interaction relation. For factuality evaluation, we incorporate templates for entities from 6 major knowledge domains commonly encountered in everyday communication, i.e., sports, politics, entertainment, religion, culture, and geography. These templates, such as "\textit{<entity1> <attribute> <relation> <entity2>}", can be instantiated to create prompts like "\textit{A toy bear in brown in front of a television}", as shown in Figure \ref{fig:overview_pipeline}. By flexibly filling templates with varied elements, our approach supports scalable and diverse data generation for hallucination evaluation. To enable this, we curate diverse and modular candidate element pools (e.g., entities, relations, attributes) that serve as the foundation for prompt generation, allowing for flexible adaptation to different template requirements. These candidate element pools are partly constructed manually and partly extracted from existing dataset labels, such as those from ImageNet~\citep{deng2009imagenet}, to enhance diversity and coverage. Specifically, for factuality-related data, we build domain-specific entity pools. For each entity, we retrieve its Wikipedia abstract via the DBpedia~\citep{auer2007dbpedia} database, extract factual claims from the retrieved abstracts, and then generate corresponding counterfactual claims by selectively modifying key information. This process results in a structured knowledge base for subsequent evaluation.

\textbf{Image Generation.} During the image generation stage, we randomly sample elements from the corresponding candidate pools and fill them into predefined type-specific templates to dynamically construct textual prompts. These prompts are then fed into an advanced text-to-image generation model, such as Stable Diffusion 3.5~\citep{stablediffusion3.5}, to synthesize the corresponding images. To ensure the quality of the generated images, we apply VQAScore~\citep{lin2024vqascore} for data filtering, and aggregate evaluation scores from multiple foundation models, such as CLIP-FlanT5~\citep{lin2024vqascore}, to further improve reliability.
Images with average scores below $\alpha$=0.85 or high inter-model variance over $\beta$=0.5 are sent for manual review. Notably, for scene text, we also incorporate professional OCR tool such as iFLYTEK OCR~\citep{iflytekocr} to verify textual content for additional quality control. We assess the final data quality through human verification on sampled data.
We observe that nearly all samples retained after the filtering process exhibit strong alignment with their generation prompts.

\textbf{Instruction Generation.} During the instruction generation stage, we design type-specific instruction templates for 4 types of questions, yes-or-no questions (YNQ), multiple-choice questions (MCQ), free-form questions (FFQ), and image captioning (IC), to support both discriminative and generative tasks. Given that the textual prompts already contain sufficient image-related information, we automatically extract relevant content to generate instructions along with corresponding ground truth answers. For example, as illustrated in Figure \ref{fig:overview_pipeline}, an instruction template like "\textit{What is the <attribute> of the <entity> in the image?}" can be instantiated as "\textit{What is the color of the teddy bear in the image?}", with the ground truth answer "\textit{Brown.}"
Questions targeting faithfulness hallucinations focus on whether the model accurately understands visual perceptual details (e.g., color, shape), while those addressing factuality hallucinations assess the model's ability to identify domain-specific entities and reason over associated factual knowledge.

\textbf{Induction Design.} To further evaluate models' susceptibility to hallucination under perturbed inputs, we design a hierarchical hallucination induction framework comprising increasingly challenging and realistic noisy scenarios, with perturbations applied at the image, instruction, and combination levels.

At the image level, we apply four representative perturbation types commonly encountered in real-world scenarios: style transformation, image corruption, adversarial noise, and scene text injection. Specifically, style transformation randomly selects a visual style (e.g., watercolor, sketch) from a predefined list and incorporates it into the prompt to produce stylized images. Image corruption simulates degraded visual quality by randomly applying operations such as salt-and-pepper noise, Gaussian blur, or JPEG compression. Adversarial noise is conducted under a black-box attack setting that utilizes EVA-CLIP ViT-G~\citep{sun2023evaclip} as a proxy visual encoder and applies untargeted PGD~\citep{madry2017pgd} attack to generate adversarial images. Scene text insertion places distractor text within the image to disrupt the model's visual understanding.

\begin{table*}[t]
\caption{Evaluation results under discriminative tasks across clean and noisy scenarios. -P denotes combination with misleading prefix, and RR denotes average Resistance Rate. Top-2 results are bolded and underlined, respectively.}
\vspace{-10 pt}
\begin{adjustbox}{max width=\linewidth}
\begin{tabular}{lccccccccccccc}
\hline
\multicolumn{1}{c}{} & \multicolumn{1}{c}{} & \multicolumn{1}{c}{} & \multicolumn{4}{c}{\textbf{Image}}          & \multicolumn{2}{c}{\textbf{Instruction}}   & \multicolumn{4}{c}{\textbf{Combination}}   & \multicolumn{1}{c}{}                   \\ \cmidrule(lr){4-7} \cmidrule(lr){8-9} \cmidrule(lr){10-13}  
\multicolumn{1}{l}{\multirow{-2}{*}{\textbf{Model}}} & \multicolumn{1}{c}{\multirow{-2}{*}{\textbf{Blind}}} & \multicolumn{1}{c}{\multirow{-2}{*}{\textbf{Clean}}} & \textbf{STY}                           & \textbf{CRP}                           & \textbf{ADV} & \textbf{TXT} & \textbf{SYN} & \textbf{PFX}                           & \textbf{STY-P}                         & \textbf{CRP-P}                         & \textbf{ADV-P}                         & \textbf{TXT-P}      & \multicolumn{1}{c}{\multirow{-2}{*}{\textbf{RR}}}                  \\ \hline
Shikra-7B~\citep{chen2023shikra}                                   & 0.335                                                & \cellcolor[HTML]{FAFDFB}0.681                        & \cellcolor[HTML]{FEF5F5}0.640          & \cellcolor[HTML]{FEFDFD}0.664          & \cellcolor[HTML]{FEF8F8}0.648          & \cellcolor[HTML]{FCC1C2}0.479          & \cellcolor[HTML]{FEEAEB}0.607          & \cellcolor[HTML]{FBAEAF}0.418          & \cellcolor[HTML]{FA9D9E}0.365          & \cellcolor[HTML]{FBA9AA}0.404          & \cellcolor[HTML]{FA999A}0.353          & \cellcolor[HTML]{F98687}0.294          & \cellcolor[HTML]{FEEFEF}0.621          \\
InstructBLIP-Vicuna-13B~\citep{dai2024instructblip}                              & 0.417                                                & \cellcolor[HTML]{F0F9F2}0.698                        & \cellcolor[HTML]{F4FBF6}0.691          & \cellcolor[HTML]{F4FBF6}0.690          & \cellcolor[HTML]{FEF1F2}0.630          & \cellcolor[HTML]{FEFDFD}0.667          & \cellcolor[HTML]{DCF1E1}0.734          & \cellcolor[HTML]{FEF5F5}0.641          & \cellcolor[HTML]{FEF3F3}0.634          & \cellcolor[HTML]{FEF3F3}0.635          & \cellcolor[HTML]{FEEEEE}0.620          & \cellcolor[HTML]{FEEDED}0.615          & \cellcolor[HTML]{96D4A7}{\ul 0.857}    \\
Otter~\citep{li2023otter}                                                & 0.419                                                & \cellcolor[HTML]{F0F9F2}0.698                        & \cellcolor[HTML]{F7FCF8}0.685          & \cellcolor[HTML]{F6FCF8}0.686          & \cellcolor[HTML]{FEFBFB}0.660          & \cellcolor[HTML]{FBB8B9}0.450          & \cellcolor[HTML]{F9FDFA}0.681          & \cellcolor[HTML]{FAA5A7}0.392          & \cellcolor[HTML]{FA9E9F}0.370          & \cellcolor[HTML]{FA9B9D}0.361          & \cellcolor[HTML]{FA9799}0.348          & \cellcolor[HTML]{F86C6E}0.212          & \cellcolor[HTML]{FDE8E8}0.599          \\
InstructBLIP-Vicuna-7B~\citep{dai2024instructblip}                      & 0.402                                                & \cellcolor[HTML]{E6F5EA}0.715                        & \cellcolor[HTML]{EAF7ED}0.709          & \cellcolor[HTML]{EBF7EF}0.706          & \cellcolor[HTML]{FEFAFA}0.655          & \cellcolor[HTML]{FDE5E5}0.591          & \cellcolor[HTML]{E4F4E8}0.720          & \cellcolor[HTML]{FCD0D1}0.526          & \cellcolor[HTML]{FCC7C8}0.498          & \cellcolor[HTML]{FCCDCE}0.517          & \cellcolor[HTML]{FBB1B2}0.430          & \cellcolor[HTML]{FAA4A6}0.389          & \cellcolor[HTML]{E9F6EC}0.710          \\
LLaVA-1.5-7B~\citep{liu2023llava}                                         & 0.371                                                & \cellcolor[HTML]{BEE4C8}0.788                        & \cellcolor[HTML]{C3E6CC}0.778          & \cellcolor[HTML]{BEE4C8}0.787          & \cellcolor[HTML]{C9E9D1}0.767          & \cellcolor[HTML]{FDE3E3}0.585          & \cellcolor[HTML]{D7EFDE}0.742          & \cellcolor[HTML]{FA9E9F}0.369          & \cellcolor[HTML]{FA9899}0.350          & \cellcolor[HTML]{FA9B9C}0.359          & \cellcolor[HTML]{FA9C9E}0.364          & \cellcolor[HTML]{F8797A}0.253          & \cellcolor[HTML]{FDE5E6}0.591          \\
mPLUG-Owl2~\citep{ye2024mplug}                                           & 0.397                                                & \cellcolor[HTML]{AFDEBB}0.814                        & \cellcolor[HTML]{AEDDBA}0.816          & \cellcolor[HTML]{A7DBB5}0.828          & \cellcolor[HTML]{B4E0BF}0.805          & \cellcolor[HTML]{FCC7C8}0.497          & \cellcolor[HTML]{CBE9D3}0.764          & \cellcolor[HTML]{FA999B}0.354          & \cellcolor[HTML]{FA9596}0.341          & \cellcolor[HTML]{FA9C9D}0.363          & \cellcolor[HTML]{F98F91}0.323          & \cellcolor[HTML]{F8696B}0.204          & \cellcolor[HTML]{FDDADA}0.555          \\
LLaVA-1.5-13B~\citep{liu2023llava}                               & 0.395                                                & \cellcolor[HTML]{A1D8B0}0.838                        & \cellcolor[HTML]{A7DBB5}0.828          & \cellcolor[HTML]{A7DAB4}0.828          & \cellcolor[HTML]{B1DFBD}0.810          & \cellcolor[HTML]{FDDADA}0.556          & \cellcolor[HTML]{C1E6CB}0.781          & \cellcolor[HTML]{FA9899}0.349          & \cellcolor[HTML]{F99294}0.333          & \cellcolor[HTML]{F99395}0.336          & \cellcolor[HTML]{F9898A}0.303          & \cellcolor[HTML]{F87172}0.228          & \cellcolor[HTML]{FDD9D9}0.552          \\
Qwen-VL-7B~\citep{Qwen-VL}                                           & 0.429                                                & \cellcolor[HTML]{9DD7AC}0.845                        & \cellcolor[HTML]{A0D8AF}0.840          & \cellcolor[HTML]{9FD7AE}0.842          & \cellcolor[HTML]{ADDDB9}0.818          & \cellcolor[HTML]{FDD9DA}0.555          & \cellcolor[HTML]{CEEBD6}0.758          & \cellcolor[HTML]{FA9798}0.346          & \cellcolor[HTML]{FA9496}0.339          & \cellcolor[HTML]{FA9496}0.339          & \cellcolor[HTML]{FA9698}0.345          & \cellcolor[HTML]{F86B6D}0.209          & \cellcolor[HTML]{FEEDED}0.615          \\
InstructBLIP-Flan-T5-XL~\citep{dai2024instructblip}                              & 0.452                                                & \cellcolor[HTML]{97D4A7}0.855                        & \cellcolor[HTML]{97D4A7}0.856          & \cellcolor[HTML]{99D5A8}0.853          & \cellcolor[HTML]{B8E2C3}0.798          & \cellcolor[HTML]{FCCCCC}0.512          & \cellcolor[HTML]{C2E6CB}0.780          & \cellcolor[HTML]{FA9C9D}0.362          & \cellcolor[HTML]{FA9596}0.340          & \cellcolor[HTML]{FA9B9D}0.361          & \cellcolor[HTML]{F98486}0.289          & \cellcolor[HTML]{F87D7F}0.267          & \cellcolor[HTML]{FDDEDE}0.568          \\
InstructBLIP-Flan-T5-XXL~\citep{dai2024instructblip}                             & 0.401                                                & \cellcolor[HTML]{97D4A7}0.855                        & \cellcolor[HTML]{98D5A8}0.854          & \cellcolor[HTML]{99D5A8}0.853          & \cellcolor[HTML]{C4E7CD}0.776          & \cellcolor[HTML]{FDE2E2}0.580          & \cellcolor[HTML]{A6DAB4}0.829          & \cellcolor[HTML]{FBB4B5}0.439          & \cellcolor[HTML]{FAA1A3}0.380          & \cellcolor[HTML]{FBAFB0}0.423          & \cellcolor[HTML]{F98C8E}0.314          & \cellcolor[HTML]{F9898A}0.303          & \cellcolor[HTML]{FDE7E7}0.596          \\
InternLM-XComposer-VL-7B~\citep{zhang2023internlm}                             & 0.442                                                & \cellcolor[HTML]{94D3A5}0.861                        & \cellcolor[HTML]{97D4A7}0.856          & \cellcolor[HTML]{91D2A2}0.867          & \cellcolor[HTML]{BEE4C8}0.788          & \cellcolor[HTML]{FEF6F6}0.645          & \cellcolor[HTML]{BEE4C8}0.788          & \cellcolor[HTML]{FBB1B2}0.430          & \cellcolor[HTML]{FAA6A8}0.396          & \cellcolor[HTML]{FBADAE}0.417          & \cellcolor[HTML]{F87D7E}0.265          & \cellcolor[HTML]{F98586}0.290          & \cellcolor[HTML]{FEEAEA}0.607          \\
InternVL2-8B~\citep{chen2024internvl}                                          & 0.402                                                & \cellcolor[HTML]{92D2A3}0.865                        & \cellcolor[HTML]{99D5A9}0.853          & \cellcolor[HTML]{95D3A6}0.859          & \cellcolor[HTML]{A3D9B2}0.834          & \cellcolor[HTML]{FEF8F8}0.649          & \cellcolor[HTML]{BFE5C9}0.784          & \cellcolor[HTML]{FDDEDE}0.568          & \cellcolor[HTML]{FCCFD0}0.522          & \cellcolor[HTML]{FDDDDD}0.565          & \cellcolor[HTML]{FDD4D5}0.539          & \cellcolor[HTML]{FBAFB1}0.424          & \cellcolor[HTML]{FEFEFE}0.669          \\
Phi-3-Vision~\citep{abdin2024phi3v}                                         & 0.393                                                & \cellcolor[HTML]{87CD9A}0.884                        & \cellcolor[HTML]{8ED0A0}0.872          & \cellcolor[HTML]{8CCF9E}0.876          & \cellcolor[HTML]{99D5A9}0.852          & \cellcolor[HTML]{FCD1D2}0.530          & \cellcolor[HTML]{BDE4C7}0.788          & \cellcolor[HTML]{FA9C9D}0.363          & \cellcolor[HTML]{FA989A}0.352          & \cellcolor[HTML]{FA9798}0.347          & \cellcolor[HTML]{F98D8F}0.317          & \cellcolor[HTML]{F87375}0.235          & \cellcolor[HTML]{FDE5E6}0.592          \\
MiniCPM-Llama2-V2.5~\citep{yao2024minicpm}                                  & 0.431                                                & \cellcolor[HTML]{78C78C}0.912                        & \cellcolor[HTML]{80CA94}0.897          & \cellcolor[HTML]{7ECA92}0.901          & \cellcolor[HTML]{87CD99}0.885          & \cellcolor[HTML]{FDE2E2}0.580          & \cellcolor[HTML]{A0D8AF}0.840          & \cellcolor[HTML]{FEF6F6}0.643          & \cellcolor[HTML]{FDE5E5}0.591          & \cellcolor[HTML]{FEEAEA}0.606          & \cellcolor[HTML]{FDE1E2}0.579          & \cellcolor[HTML]{FBAEAF}0.418          & \cellcolor[HTML]{E8F6EC}0.712          \\
DeepSeek-VL2~\citep{wu2024deepseek}                                         & 0.392                                                & \cellcolor[HTML]{73C589}0.921                        & \cellcolor[HTML]{6FC385}0.928          & \cellcolor[HTML]{70C486}0.925          & \cellcolor[HTML]{77C78C}0.914          & \cellcolor[HTML]{FEECEC}0.613          & \cellcolor[HTML]{93D2A4}0.864          & \cellcolor[HTML]{FDE7E7}0.597          & \cellcolor[HTML]{FDE0E0}0.575          & \cellcolor[HTML]{FDE8E8}0.600          & \cellcolor[HTML]{FDD8D8}0.550          & \cellcolor[HTML]{FBAFB0}0.421          & \cellcolor[HTML]{EDF8F0}0.703          \\
InternLM-XComposer2-VL-7B~\citep{dong2024internlm2}                            & 0.399                                                & \cellcolor[HTML]{72C588}0.922                        & \cellcolor[HTML]{78C78D}0.912          & \cellcolor[HTML]{74C68A}0.918          & \cellcolor[HTML]{7CC990}0.905          & \cellcolor[HTML]{FEFEFE}0.668          & \cellcolor[HTML]{9CD6AC}0.847          & \cellcolor[HTML]{FA9899}0.349          & \cellcolor[HTML]{F99091}0.325          & \cellcolor[HTML]{F99192}0.328          & \cellcolor[HTML]{F87779}0.248          & \cellcolor[HTML]{F8696B}0.202          & \cellcolor[HTML]{FDE4E4}0.586          \\
GLM-4V-9B~\citep{glm2024chatglm4v}                                            & 0.424                                                & \cellcolor[HTML]{71C487}0.924                        & \cellcolor[HTML]{74C58A}0.918          & \cellcolor[HTML]{73C589}0.920          & \cellcolor[HTML]{80CA93}0.898          & \cellcolor[HTML]{FEF8F8}0.649          & \cellcolor[HTML]{98D4A8}0.854          & \cellcolor[HTML]{FDDADA}0.556          & \cellcolor[HTML]{FCCACA}0.506          & \cellcolor[HTML]{FCCDCE}0.517          & \cellcolor[HTML]{FBB5B6}0.441          & \cellcolor[HTML]{F99294}0.333          & \cellcolor[HTML]{F7FCF9}0.685          \\

InternVL3-8B~\citep{zhu2025internvl3}                                         & 0.399                                                & \cellcolor[HTML]{6EC385}0.928                        & \cellcolor[HTML]{74C68A}0.918          & \cellcolor[HTML]{72C488}0.923          & \cellcolor[HTML]{76C68B}0.915          & \cellcolor[HTML]{FEF4F5}0.639          & \cellcolor[HTML]{96D4A7}0.857          & \cellcolor[HTML]{FFFFFF}0.671          & \cellcolor[HTML]{FEF2F2}0.631          & \cellcolor[HTML]{FEF8F8}0.649          & \cellcolor[HTML]{FEEDED}0.616          & \cellcolor[HTML]{FDD7D7}0.546          & \cellcolor[HTML]{D0ECD7}0.755          \\
Gemini-2.0-flash~\citep{gemini2.0flash}                                     & 0.308                                                & \cellcolor[HTML]{6BC282}0.935                        & \cellcolor[HTML]{6CC283}{\ul 0.933}    & \cellcolor[HTML]{68C07F}{\ul 0.940}    & \cellcolor[HTML]{70C486}{\ul 0.925}    & \cellcolor[HTML]{BCE3C6}\textbf{0.791} & \cellcolor[HTML]{8BCF9D}\textbf{0.878} & \cellcolor[HTML]{A1D8B0}\textbf{0.838} & \cellcolor[HTML]{AADCB7}\textbf{0.823} & \cellcolor[HTML]{A3D9B2}\textbf{0.834} & \cellcolor[HTML]{BAE3C5}\textbf{0.794} & \cellcolor[HTML]{EAF7EE}\textbf{0.708} & \cellcolor[HTML]{90D1A1}\textbf{0.869} \\
InternVL3-14B~\citep{zhu2025internvl3}                                        & 0.406                                                & \cellcolor[HTML]{68C07F}{\ul 0.941}                  & \cellcolor[HTML]{6EC384}0.930          & \cellcolor[HTML]{6BC282}0.935          & \cellcolor[HTML]{70C486}\textbf{0.926} & \cellcolor[HTML]{FEF6F6}0.644          & \cellcolor[HTML]{93D2A4}0.863          & \cellcolor[HTML]{E0F2E5}0.727          & \cellcolor[HTML]{ECF7EF}0.705          & \cellcolor[HTML]{EBF7EF}0.706          & \cellcolor[HTML]{FEF3F3}0.635          & \cellcolor[HTML]{FDDADA}0.555          & \cellcolor[HTML]{C4E7CE}0.775          \\
Qwen-VL-Max~\citep{qwenvlmax}                                          & 0.379                                                & \cellcolor[HTML]{64BF7C}\textbf{0.947}               & \cellcolor[HTML]{6AC181}\textbf{0.937} & \cellcolor[HTML]{63BE7B}\textbf{0.948} & \cellcolor[HTML]{72C487}0.923          & \cellcolor[HTML]{E4F4E9}{\ul 0.719}    & \cellcolor[HTML]{8CCF9E}{\ul 0.876}    & \cellcolor[HTML]{C1E6CB}0.781          & \cellcolor[HTML]{DBF0E1}0.735          & \cellcolor[HTML]{D3EDD9}0.750          & \cellcolor[HTML]{FCFEFC}0.677          & \cellcolor[HTML]{FEF6F6}{\ul 0.645}    & \cellcolor[HTML]{AEDDBA}0.816         
\\
Average                   & 0.400 & \cellcolor[HTML]{9AD5A9}0.849          & \cellcolor[HTML]{9ED7AD}0.840          & \cellcolor[HTML]{9CD6AB}0.845          & \cellcolor[HTML]{ACDDB9}0.816          & \cellcolor[HTML]{FEEAEA}0.600          & \cellcolor[HTML]{B9E2C3}0.792          & \cellcolor[HTML]{FCCDCE}0.510          & \cellcolor[HTML]{FCC4C4}0.481          & \cellcolor[HTML]{FCC8C9}0.496          & \cellcolor[HTML]{FBB8B9}0.445          & \cellcolor[HTML]{FA9FA0}0.369          & \cellcolor[HTML]{FCFEFD}0.668         \\

\hline
\end{tabular}
\end{adjustbox}
\label{tab:dis}
\vspace{-5 pt}
\end{table*}

\begin{table}[t]
\caption{Evaluation results under generative tasks across clean and noisy scenarios. RR denotes average Resistance Rate. Top-2 results are bolded and underlined, respectively.}
\vspace{-10 pt}
\begin{adjustbox}{max width=1\linewidth}

\begin{tabular}{lcccccc}
\hline
\multicolumn{1}{c}{} & \multicolumn{1}{c}{}                               & \multicolumn{4}{c}{\textbf{Image}}             & \multicolumn{1}{c}{}                   \\ \cmidrule(lr){3-6} 
\multicolumn{1}{l}{\multirow{-2}{*}{\textbf{Model}}} & \multicolumn{1}{c}{\multirow{-2}{*}{\textbf{Clean}}} & \textbf{STY}                           & \textbf{CRP}                           & \textbf{ADV}                           & \textbf{TXT}     & \multicolumn{1}{c}{\multirow{-2}{*}{\textbf{RR}}}                       \\ \hline
Otter~\citep{li2023otter} & \cellcolor[HTML]{FAA2A3}0.448          & \cellcolor[HTML]{FA9A9B}0.429          & \cellcolor[HTML]{FA9799}0.423          & \cellcolor[HTML]{F99192}0.407          & \cellcolor[HTML]{F87375}0.337          & \cellcolor[HTML]{FCC4C4}0.527          \\
InternLM-XComposer-VL-7B~\citep{zhang2023internlm}& \cellcolor[HTML]{FCC3C4}0.526          & \cellcolor[HTML]{FBB2B3}0.486          & \cellcolor[HTML]{FCBEBF}0.515          & \cellcolor[HTML]{F8696B}0.312          & \cellcolor[HTML]{F99193}0.408          & \cellcolor[HTML]{FCCCCC}0.546          \\
Shikra-7B~\citep{chen2023shikra}& \cellcolor[HTML]{FCCFD0}0.555          & \cellcolor[HTML]{FBB9BA}0.502          & \cellcolor[HTML]{FCC4C4}0.527          & \cellcolor[HTML]{F99293}0.410          & \cellcolor[HTML]{FAA8AA}0.463          & \cellcolor[HTML]{FDDBDC}0.583          \\
LLaVA-1.5-7B~\citep{liu2023llava}& \cellcolor[HTML]{FDD6D7}0.571          & \cellcolor[HTML]{FCCDCE}0.549          & \cellcolor[HTML]{FDD4D5}0.566          & \cellcolor[HTML]{FCCCCC}0.546          & \cellcolor[HTML]{FBADAE}0.474          & \cellcolor[HTML]{FEFFFE}0.668          \\
LLaVA-1.5-13B~\citep{liu2023llava}& \cellcolor[HTML]{FEEBEB}0.619          & \cellcolor[HTML]{FDE0E1}0.594          & \cellcolor[HTML]{FEE9EA}0.616          & \cellcolor[HTML]{FDD8D9}0.576          & \cellcolor[HTML]{FBB5B5}0.491          & \cellcolor[HTML]{FCFEFC}0.671          \\
InterVL2-8B~\citep{chen2024internvl}& \cellcolor[HTML]{FEF4F4}0.641          & \cellcolor[HTML]{FEF0F0}0.631          & \cellcolor[HTML]{FDE8E9}0.614          & \cellcolor[HTML]{FCC8C9}0.538          & \cellcolor[HTML]{FA9D9F}0.437          & \cellcolor[HTML]{FEF3F3}0.639          \\
Phi-3-Vision~\citep{abdin2024phi3v}& \cellcolor[HTML]{FEFAFA}0.656          & \cellcolor[HTML]{F2FAF4}0.683          & \cellcolor[HTML]{FDE6E6}0.608          & \cellcolor[HTML]{FA9395}0.414          & \cellcolor[HTML]{FBB4B5}0.490          & \cellcolor[HTML]{F6FBF7}0.678          \\
mPLUG-Owl2~\citep{ye2024mplug}& \cellcolor[HTML]{FEFCFC}0.660          & \cellcolor[HTML]{FEEBEC}0.621          & \cellcolor[HTML]{FEECEC}0.623          & \cellcolor[HTML]{FDDADB}0.581          & \cellcolor[HTML]{FBBBBC}0.507          & \cellcolor[HTML]{FAFDFB}0.673          \\
InstructBLIP-Flan-T5-XL~\citep{dai2024instructblip}& \cellcolor[HTML]{F5FBF7}0.679          & \cellcolor[HTML]{FEEBEB}0.619          & \cellcolor[HTML]{FEF8F8}0.651          & \cellcolor[HTML]{FA9496}0.415          & \cellcolor[HTML]{FBBCBD}0.509          & \cellcolor[HTML]{F0F9F2}0.686          \\
InstructBLIP-Vicuna-13B~\citep{dai2024instructblip}& \cellcolor[HTML]{E3F4E8}0.701          & \cellcolor[HTML]{FEFEFE}0.664          & \cellcolor[HTML]{FCFEFD}0.670          & \cellcolor[HTML]{FBAEAF}0.476          & \cellcolor[HTML]{FDDBDB}0.582          & \cellcolor[HTML]{DBF0E1}0.711          \\
InstructBLIP-Flan-T5-XXL~\citep{dai2024instructblip}& \cellcolor[HTML]{CEEBD6}0.728          & \cellcolor[HTML]{EDF8F0}0.689          & \cellcolor[HTML]{DBF0E0}0.712          & \cellcolor[HTML]{FAA7A8}0.459          & \cellcolor[HTML]{FCC0C1}0.518          & \cellcolor[HTML]{EBF7EE}0.692          \\
InstructBLIP-Vicuna-7B~\citep{dai2024instructblip}& \cellcolor[HTML]{CDEAD5}0.729          & \cellcolor[HTML]{F0F9F3}0.685          & \cellcolor[HTML]{E8F6EB}0.696          & \cellcolor[HTML]{FAA8A9}0.462          & \cellcolor[HTML]{FDE0E1}0.595          & \cellcolor[HTML]{D3EDDA}0.722          \\
InternVL3-8B~\citep{zhu2025internvl3}& \cellcolor[HTML]{C6E7CE}0.739          & \cellcolor[HTML]{DAF0DF}0.714          & \cellcolor[HTML]{CEEBD6}0.728          & \cellcolor[HTML]{CFEBD6}0.727          & \cellcolor[HTML]{FCCCCD}0.548          & \cellcolor[HTML]{B5E0C0}0.759          \\
InternVL3-14B~\citep{zhu2025internvl3}& \cellcolor[HTML]{C0E5CA}0.746          & \cellcolor[HTML]{CAE9D2}0.733          & \cellcolor[HTML]{D0ECD8}0.725          & \cellcolor[HTML]{FEF4F4}0.641          & \cellcolor[HTML]{FCC8C9}0.539          & \cellcolor[HTML]{C0E5CA}0.746          \\
InternLM-XComposer2-VL-7B~\citep{dong2024internlm2}& \cellcolor[HTML]{B2DFBE}0.764          & \cellcolor[HTML]{DAF0DF}0.714          & \cellcolor[HTML]{BBE3C6}0.751          & \cellcolor[HTML]{CAE9D2}0.734          & \cellcolor[HTML]{FDDCDD}0.585          & \cellcolor[HTML]{9DD7AC}0.789          \\
Qwen-VL-7b~\citep{Qwen-VL}& \cellcolor[HTML]{ACDDB9}0.771          & \cellcolor[HTML]{BFE5C9}0.747          & \cellcolor[HTML]{C5E7CE}0.739          & \cellcolor[HTML]{E4F4E9}0.700          & \cellcolor[HTML]{FDDFE0}0.592          & \cellcolor[HTML]{A6DAB4}0.778          \\
GLM-4V-9B~\citep{glm2024chatglm4v}& \cellcolor[HTML]{A5DAB3}0.779          & \cellcolor[HTML]{B8E2C3}0.755          & \cellcolor[HTML]{AFDEBB}0.767          & \cellcolor[HTML]{ECF7EF}0.691          & \cellcolor[HTML]{FDDBDB}0.582          & \cellcolor[HTML]{A1D8B0}0.784          \\
MiniCPM-Llama2-V2.5~\citep{yao2024minicpm}& \cellcolor[HTML]{A2D8B0}0.784          & \cellcolor[HTML]{BBE3C5}0.752          & \cellcolor[HTML]{AADCB7}0.773          & \cellcolor[HTML]{C9E9D1}0.734          & \cellcolor[HTML]{FDD5D6}0.569          & \cellcolor[HTML]{A4D9B2}0.780          \\
DeepSeek-VL2~\citep{wu2024deepseek}& \cellcolor[HTML]{7CC990}0.831          & \cellcolor[HTML]{84CC97}{\ul 0.821}    & \cellcolor[HTML]{7FCA93}{\ul 0.826}    & \cellcolor[HTML]{9FD7AE}{\ul 0.786}    & \cellcolor[HTML]{FDE2E2}0.599          & \cellcolor[HTML]{76C68B}{\ul 0.838}    \\
Qwen-VL-Max~\citep{qwenvlmax}& \cellcolor[HTML]{72C588}{\ul 0.843}    & \cellcolor[HTML]{83CC96}\textbf{0.822} & \cellcolor[HTML]{8DD09F}0.809          & \cellcolor[HTML]{CBE9D3}0.732          & \cellcolor[HTML]{FEFEFE}\textbf{0.664} & \cellcolor[HTML]{89CE9B}0.815          \\
Gemini-2.0-flash~\citep{gemini2.0flash}& \cellcolor[HTML]{6CC283}\textbf{0.851} & \cellcolor[HTML]{84CC97}{\ul 0.821}    & \cellcolor[HTML]{63BE7B}\textbf{0.861} & \cellcolor[HTML]{92D2A3}\textbf{0.804} & \cellcolor[HTML]{FEF1F2}{\ul 0.635}    & \cellcolor[HTML]{72C487}\textbf{0.844} \\
Average                   & \cellcolor[HTML]{E7F5EB}0.696          & \cellcolor[HTML]{FEFFFE}0.668          & \cellcolor[HTML]{F7FCF9}0.676          & \cellcolor[HTML]{FDD9DA}0.578          & \cellcolor[HTML]{FCC5C6}0.530          & \cellcolor[HTML]{DCF1E1}0.711  \\       
\hline
\end{tabular}

\end{adjustbox}
\vspace{-10 pt}
\label{tab:gen}
\end{table}

At the instruction level, we introduce linguistic perturbations through confusing synonyms and misleading prefixes. The former replaces distractor options in YNQ and MCQ questions with synonyms that are semantically closer to the correct answer, thereby increasing the likelihood of model misinterpretation. For example, given the ground truth answer "\textit{Raccoon}", distractors originally randomly sampled as ["\textit{Leopard}", "\textit{Ant}", "\textit{Goose}"] are replaced with semantically closer alternatives such as ["\textit{Bear}", "\textit{Squirrel}", "\textit{Wolf}"], as illustrated in Figure \ref{fig:examples_dataset}. The latter method further adds a misleading statement derived from the distractor content as an instruction prefix, leveraging language priors to bias the model's understanding of the image. An example of such a misleading prefix is "\textit{There seems to be a squirrel in the forest}", aiming to challenge the model's resistance to linguistic priors.

At the combination level, we integrate both image and instruction perturbations to create more complex scenarios, simulating real-world settings where models are exposed to simultaneous visual and linguistic noise. This hallucination induction framework enables a hierarchical evaluation of LVLMs' stability and hallucination resistance under varying degrees and forms of perturbation.

\subsection{Evaluation Metrics}
\label{sec:metric}

For discriminative tasks, we adopt accuracy as the evaluation metric. For generative tasks, we follow previous work ~\citep{sun2023mmhal, liu2023gaive}
, employing an LLM-as-a-Judge approach to assess hallucinations. Specifically, given the image content, instruction, and ground-truth answer, we prompt an advanced LLM
to determine whether the model's response is hallucination-free, and calculate the non-hallucination rate as the evaluation metric. Compared to CHAIR~\citep{rohrbach2018chair}, which focuses solely on object-level hallucination detection, the LLM-as-a-Judge approach offers broader generalizability by supporting diverse hallucination types and task formats. Additionally, the non-hallucination rate is a comparable metric to accuracy, enabling direct derivation of an overall average performance score.

To quantify model resistance to perturbations under various hallucination-inducing noisy scenarios, we propose the Resistance Rate (RR) metric. It measures the proportion of input image-text pairs with non-hallucinated responses under clean scenario that remain non-hallucinated when subject to perturbations. For a LVLM with parameters \(\theta\), given a set of clean input image-instruction pairs \((i, t) \in I \times T\), and corresponding perturbed input pairs \((\tilde{i}, \tilde{t})\), the Resistance Rate is defined as:
\[
RR(\theta) = \frac{
\sum_{(i,t) \in I \times T} \mathbf{1}_{\text{nh}}(i, t, \theta) \cdot \mathbf{1}_{\text{nh}}(\tilde{i}, \tilde{t}, \theta)
}{
\sum_{(i,t) \in I \times T} \mathbf{1}_{\text{nh}}(i, t, \theta)
},
\]
where \(\mathbf{1}_{\text{nh}}(i, t, \theta) = 1\) if model \(\theta\) produces a non-hallucinated response for \((i, t)\), and 0 otherwise.

\begin{figure}[h]
  \centering
    \includegraphics[width=0.95\linewidth]{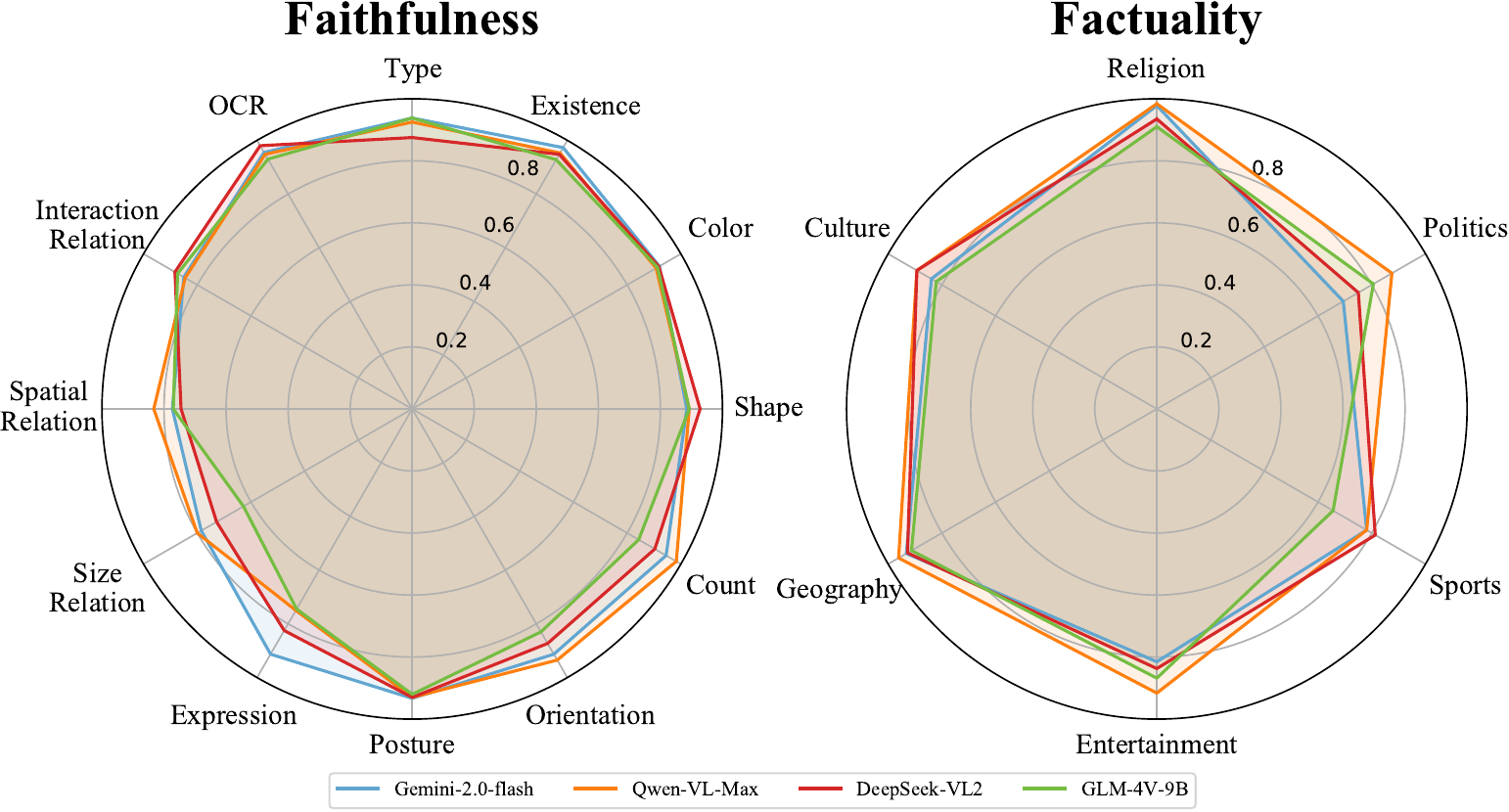}
  \vspace{-5 pt}
  \caption{Comparison of the top-4 LVLMs on fine-grained evaluation dimensions.}
  \vspace{-20 pt}
  \label{rada}
\end{figure}

\section{Results and Analysis}
\subsection{Evaluation Results}
We evaluate 19 popular open-source LVLMs on SHALE, e.g., GLM-4V-9B~\citep{glm2024chatglm4v}, and DeepSeek-VL2~\citep{wu2024deepseek}
, as well as 2 powerful closed-source models,
Gemini-2.0-Flash~\citep{gemini2.0flash}, and Qwen-VL-MAX~\citep{qwenvlmax}. For noisy scenarios, we evaluate model performance under image-level, instruction-level, and combination-level perturbations, including style transformation (STY), image corruption (CRP), adversarial noise (ADV), scene text insertion (TXT), confusing synonyms (SYN), misleading prefixes (PFX), and their combinations. Notably, for generative tasks, we leverage only image-level perturbations, as they already significantly degrade model performance. 

\textbf{Performance across Tasks.} Table~\ref{tab:dis} and Table~\ref{tab:gen} present the LVLMs' performance on discriminative and generative tasks, respectively. Overall, generative tasks are more challenging than discriminative tasks, as evidenced by a significant average performance drop of over 15\% for LVLMs in generative settings compared to discriminative ones under clean scenarios. Under noisy scenarios, LVLMs are more susceptible to performance degradation in generative tasks, while exhibiting greater robustness on discriminative tasks, highlighting the need to evaluate hallucinations on generative tasks, which may better reflect real-world challenges. Moreover, different models demonstrate varying strengths across task types. For instance, InstructBLIP-Vicuna-7B~\citep{dai2024instructblip} performs poorly on discriminative tasks but excels in generative tasks, underscoring the necessity of comprehensive evaluation across tasks.

\textbf{Performance across Fine-grained Dimensions.} For further analysis, we visualize the performance of the top-4 models across both fine-grained faithfulness and factuality hallucination dimensions, as shown in Figure~\ref{rada}. Most of them struggle more with factuality hallucinations than faithfulness hallucinations, highlighting the critical need for improving factual accuracy, especially in sports and politics domains. Regarding faithfulness hallucinations, relational hallucinations, particularly size relation and spatial relation, emerge as the most problematic dimensions.

\textbf{Performance across Hallucination-inducing Scenarios.} As shown in Table~\ref{tab:dis} and Table~\ref{tab:gen}, semantic perturbations, such as scene text injection and misleading instruction prefixes, are considerably more detrimental than low-level visual noise. At the image level, scene text injection is the most disruptive, significantly increasing hallucination rates for most LVLMs. Adversarial noise also causes great degradation, particularly in generative tasks, likely due to adversarial perturbations shifting the image representation in the high-dimensional feature space. Models like InternLM-XComposer2-VL-7B~\citep{dong2024internlm2} show strong robustness to adversarial noise, with performance drops under 3\% across both task types, likely due to the architectural gap between its visual encoder and the adversarial proxy encoder, as well as additional visual fine-tuning. At the instruction level, confusing synonyms cause minor accuracy drops, while misleading instruction prefixes have a significant negative effect, particularly on open-source models, highlighting the influence of contextual information. Some models, such as Qwen-VL-7B~\citep{Qwen-VL}, degrade to near-random guessing accuracy (i.e., 38.1\%).   Combination-level perturbations exhibit compounded challenges, with the joint use of scene text injection and misleading prefixes forming the most difficult setting. 
These findings suggest that most LVLMs are highly sensitive to linguistically or contextually misleading cues, which may not be easily mitigated by scaling and standard pretraining. 

\textbf{Performance across Models.} As shown in Table~\ref{tab:dis} and Table~\ref{tab:gen}, closed-source LVLMs consistently outperform their open-source counterparts, especially in noisy scenarios. We attribute this to the higher-quality data used during training and instruction tuning stages, as well as specific model policies explicitly designed to enhance factual accuracy~\citep{gemini2.0flash}. Some open-source models, such as InternVL3-14B~\citep{chen2024internvl} and DeepSeek-VL2~\citep{wu2024deepseek}, also perform competitively under clean scenario, reaching levels comparable to those of closed-source models, which indicates that their architectural or training design effectively contributes to hallucination mitigation. Additionally, we observe that strong performance under clean conditions does not always correlate with hallucination resistance. For example, models such as Gemini-2.0-Flash~\citep{gemini2.0flash} and InstructBLIP-Vicuna-13B~\citep{dai2024instructblip}, although not top performers on discriminative tasks in clean settings, demonstrate remarkable robustness under hallucination-inducing scenarios with high resistance rates. These findings underscore the importance of evaluating LVLMs in challenging environments, as hallucination resistance is a critical aspect of reliability but is often overlooked by standard benchmarks.

\subsection{Discussion}

\textbf{Reliability and Validity of SHALE.} To verify the effectiveness of SHALE in mitigating data leakage, we first conduct a blind-setting experiment where models are provided with only the instructions without images. As shown in Table~\ref{tab:dis}, most models achieve performance close to random guessing (i.e., 38.1\%), indicating that they have never seen or exploited spurious correlations from training data. To further assess the validity and reliability of our benchmark, we adopt the Multimodal Gain (MG) and Multimodal Leakage (ML) metrics proposed by~\citet{chen2024mmstar}. MG measures the actual benefit from multimodal input, while ML estimates data leakage from training corpora. As shown in Table~\ref{tab:ml}, SHALE achieves the lowest ML and highest MG among all benchmarks, including hallucination benchmarks like POPE~\citep{li2023pope} and general multimodal benchmarks like SEED~\citep{li2023seed}. These results highlight SHALE's effectiveness in reducing leakage risk while preserving the informativeness and challenge level of the evaluation, making it a more trustworthy and robust benchmark for hallucination assessment.

\begin{table}[t]
\caption{Evaluation results on the reliability and validity of existing benchmarks based on LLaVA.}
\vspace{-5pt}
\begin{adjustbox}{max width=1\linewidth}
\centering
\setlength{\tabcolsep}{2pt}
\begin{tabular}{lcccccccccccc}
\hline
\multicolumn{1}{l}{\multirow{2}{*}{\textbf{Model}}} & \multicolumn{2}{c}{\textbf{SHALE (Ours)}} & \multicolumn{2}{c}{\textbf{POPE}} & \multicolumn{2}{c}{\textbf{AMBER}} & \multicolumn{2}{c}{\textbf{SEED}}\\  \cmidrule(lr){2-3} \cmidrule(lr){4-5} \cmidrule(lr){6-7} \cmidrule(lr){8-9} 
\multicolumn{1}{c}{}                                & \textbf{MG↑}       & \textbf{ML↓}      & \textbf{MG↑}    & \textbf{ML↓}    & \textbf{MG↑}     & \textbf{ML↓}    & \textbf{MG↑}    & \textbf{ML↓}       \\ \hline

LLaVA-1.5-7B~\citep{liu2023llava}  & \textbf{41.25} & \textbf{0.00} & 28.43 & \textbf{0.00} & 4.03  & 6.20 & 28.1 & 4.9  \\
LLaVA-1.5-13B~\citep{liu2023llava} & \textbf{43.79} & \textbf{0.25} & 31.17 & 1.10          & 10.63 & 6.83 & 31.1 & 10.7 
\\ \hline
\end{tabular}
\label{tab:ml}
\end{adjustbox}
\vspace{-20pt}

\end{table}

\textbf{Potential Applications.}
Although SHALE is currently designed for evaluation purposes, it holds promising potential for other applications, such as hallucination mitigation. For instance, our data generation pipeline can be leveraged to create fine-tuning datasets aimed at enhancing LVLMs' robustness against hallucinations in a targeted manner. Besides, the proposed automatic data construction pipeline can be easily extended to other specialized domains.

\textbf{Potential Ethical Concerns.} SHALE builds upon an automated data construction pipeline. All images in our benchmark are generated using predefined prompt templates filled with various image elements. We have verified that these images contain no identifiable personal data or depictions of explicit violence or gore, ensuring that the benchmark poses no adverse impact on individuals or communities. Furthermore, all experiments are conducted in strict adherence to ethical guidelines.

\textbf{Limitations.}
SHALE generates images using a text-to-image diffusion model based on predefined prompt templates. While this approach ensures controllability and scalability, it may not fully capture the complexity and diversity of real-world scenarios, thereby limiting the benchmark to relatively basic evaluation settings. Moreover, SHALE currently covers only a representative subset rather than an exhaustive range of fine-grained hallucination dimensions, which calls for further expansion and refinement in future work.

\section{Conclusion}
In this paper, we propose an automated data construction pipeline and hallucination induction framework for fine-grained hallucination evaluation. The resulting SHALE benchmark supports fine-grained assessment of both faithfulness and factuality hallucinations across discriminative and generative tasks, under diverse hallucination-inducing noisy scenarios. Experiments on over 20 representative LVLMs reveal that current models still struggle with relational and factual hallucinations, and are particularly susceptible to semantically relevant perturbations. 

\section*{Acknowledgments}
This work is partially supported by Strategic Priority Research Program of the Chinese Academy of Sciences (No. XDB0680202), Beijing Nova Program (20230484368), Suzhou Frontier Technology Research Project (No. SYG202325), and Youth Innovation Promotion Association CAS.

\bibliographystyle{ACM-Reference-Format}
\bibliography{main}

\clearpage

\appendix
\section{More Construction Details}
\label{appendix:construction details}

\subsection{Data Filtering}
\label{appendix:dataquality}
To ensure the quality of the generated images, we leverage the advanced image-text alignment metric, VQAScore~\citep{lin2024vqascore}, for data filtering. Specifically, we use two strong foundation models to independently compute alignment scores and then take their average. Images with an average score below the threshold $\alpha$= 0.85 or with high inter-model variance exceeding $\beta$=0.5 are flagged for manual review. For images containing text, we further enhance filtering precision using iFLYTEK OCR~\citep{iflytekocr} to ensure accurate text recognition and alignment. 

To assess the effectiveness of our filtering pipeline, we randomly sampled 500 images from the final dataset for manual review. The evaluation criteria included: (1) whether the image content aligns with the corresponding prompt, and (2) whether the image could potentially mislead models to incorrect answers. Our review found that nearly all images demonstrated good alignment and quality. Fewer than 5\% exhibited minor issues, such as small distortions or artifacts caused by generation model limitations, which were unrelated to semantic alignment. Considering these minor imperfections are unlikely to affect evaluation outcomes, and in light of known limitations in widely-used datasets, e.g., at least 2,916 labeling errors in the ImageNet validation set~\citep{zhang2024dysca}, we consider the quality of our dataset acceptable for rigorous hallucination evaluation.

\subsection{Instruction Templates}
\label{appendix:instructiontemplate}
We design type-specific instruction templates for four question formats: yes-or-no questions (YNQ), multiple-choice questions (MCQ), free-form questions (FFQ), and image captioning (IC), supporting both discriminative and generative tasks.

Specifically, for YNQ, we construct questions that generate "yes" or "no" answers, such as "\textit{Is the <entity> <attribute> in the image?}", "\textit{Is the <entity1> <relation> <entity2> in the image?}", "\textit{Is the <fact> about <entity> true in the image?}". For MCQ, we randomly shuffle the ground-truth answer with three distractors drawn from the same candidate element set. Example templates include "\textit{What is the <attribute> of the <entity> in the image? <Options>}", and "\textit{What is the <fact> about the <entity> in the image? <Options>}". For FFQ, we simply remove the options from the MCQ templates to form open-ended questions. For IC, the instruction is a straightforward prompt, "\textit{Please describe the image.}"

\subsection{Evaluation Details}
\label{appendix:evaluationprompt}

\begin{figure}[h]
    \centering
\includegraphics[width=\linewidth]{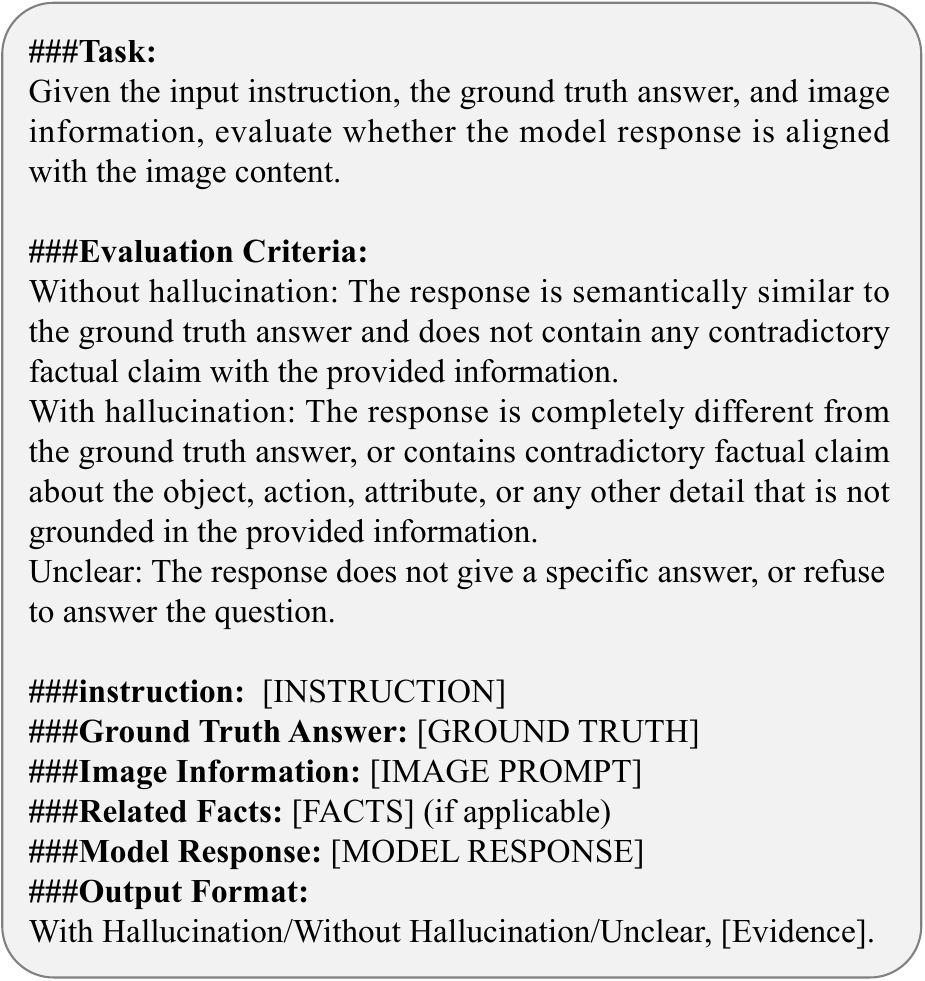}
    \caption{Evaluation prompt template for generative task.}
    \label{fig:prompt}
\end{figure}

For discriminative tasks, we adopt accuracy as the evaluation metric by directly extracting and matching the corresponding options from model responses. For generative tasks, we follow the LLM-as-a-Judge protocol in prior works~\citep{sun2023mmhal, liu2023gaive}, i.e., given the image content, instruction, and ground-truth answer, we prompt external LLM to determine whether the model's response is hallucination-free. The non-hallucination rate is then used as the evaluation metric. The detailed evaluation prompt template is illustrated in Figure \ref{fig:prompt}.

\begin{figure}[h]
    \centering
\includegraphics[width=\linewidth]{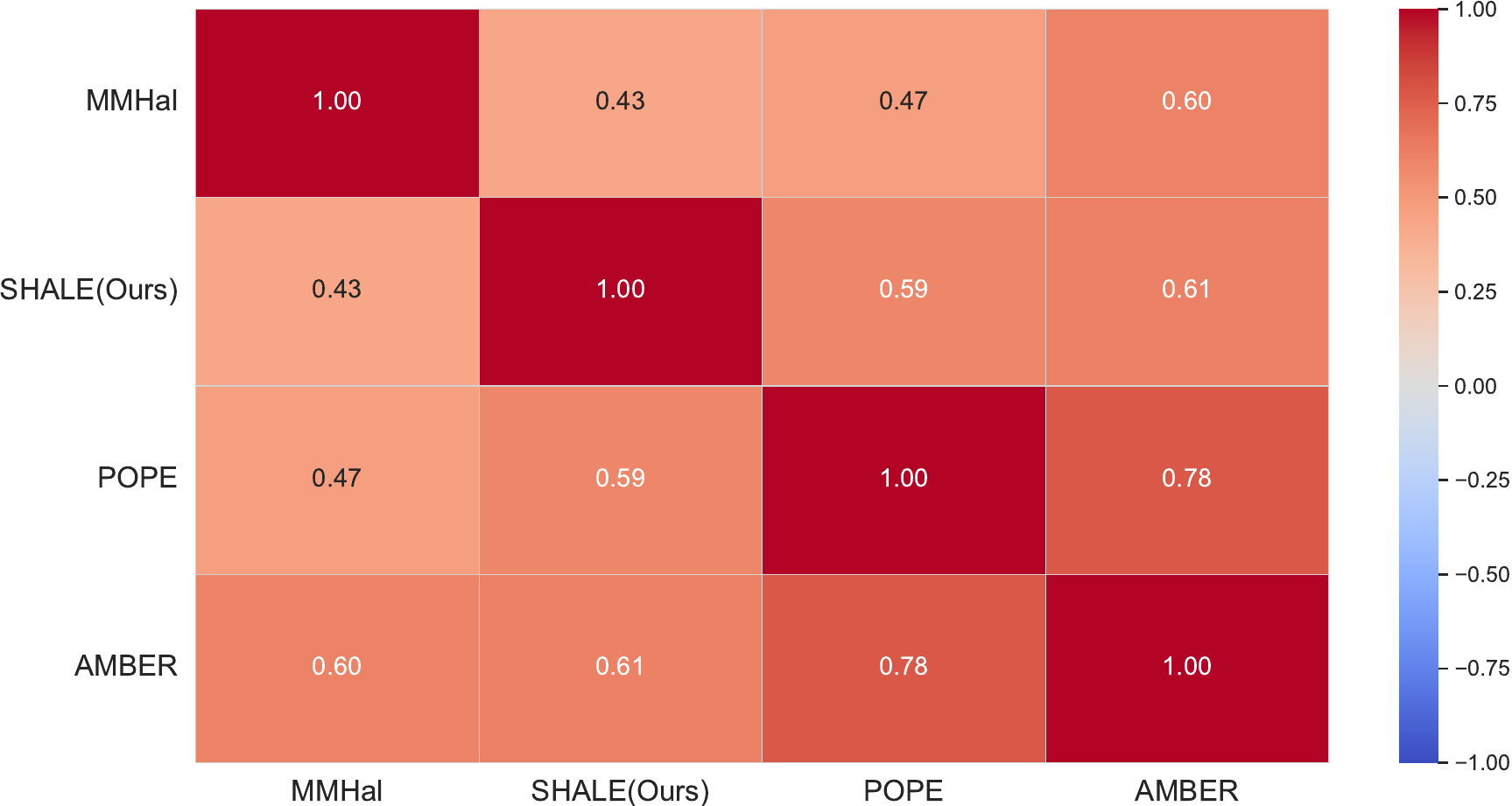}
    \caption{Visualization of the correlation between hallucination benchmark evaluation results across different LVLMs.}
    \label{fig:correlation}
\end{figure}

\begin{figure*}[h]
\subfigure{
\begin{minipage}[t]{1.0\linewidth}
\begin{center}
\includegraphics[width=0.95\linewidth]{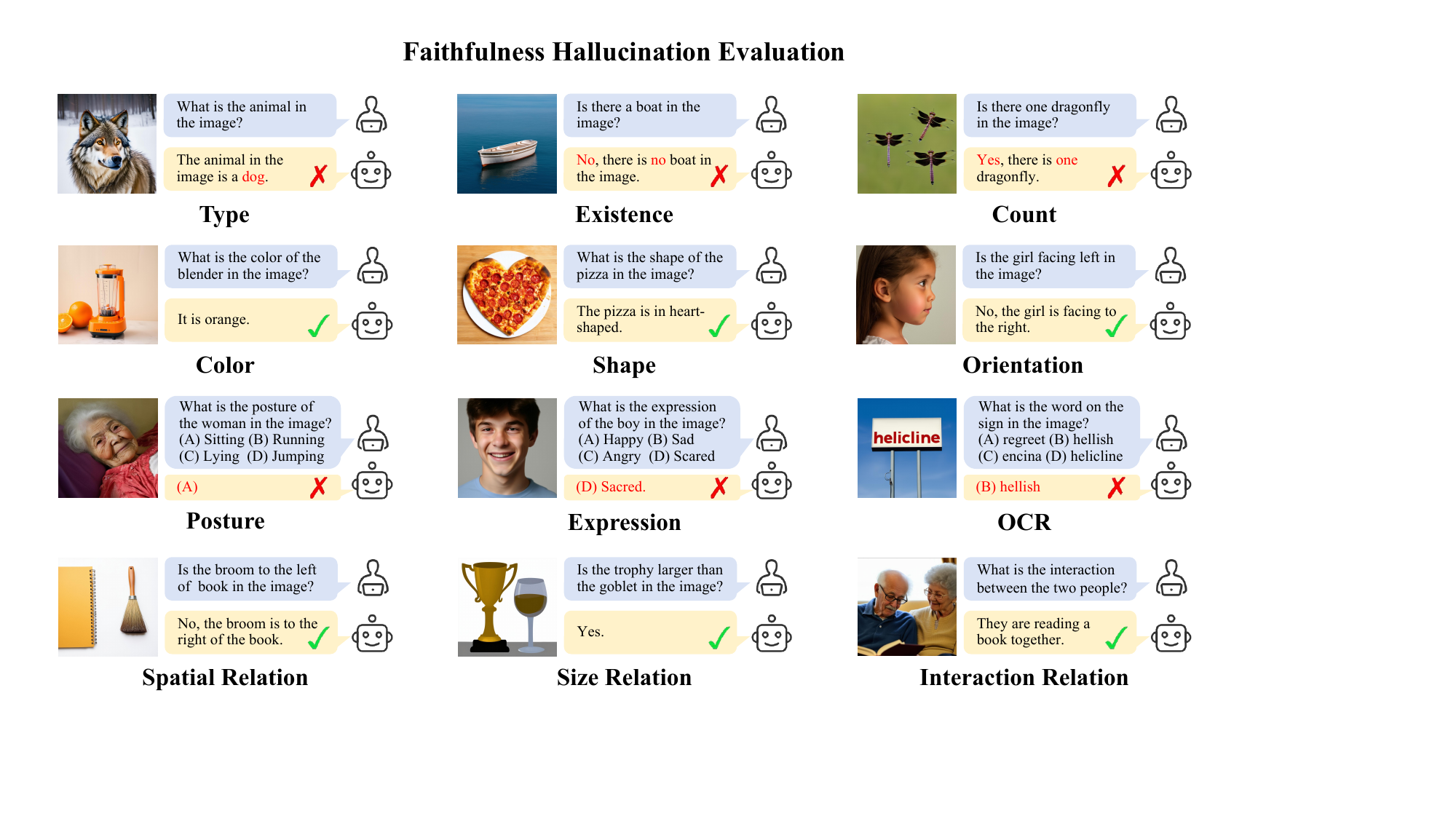}
\end{center}
\end{minipage}
}

\subfigure{
\begin{minipage}[t]{1.0\linewidth}
\begin{center}
\includegraphics[width=0.95\linewidth]{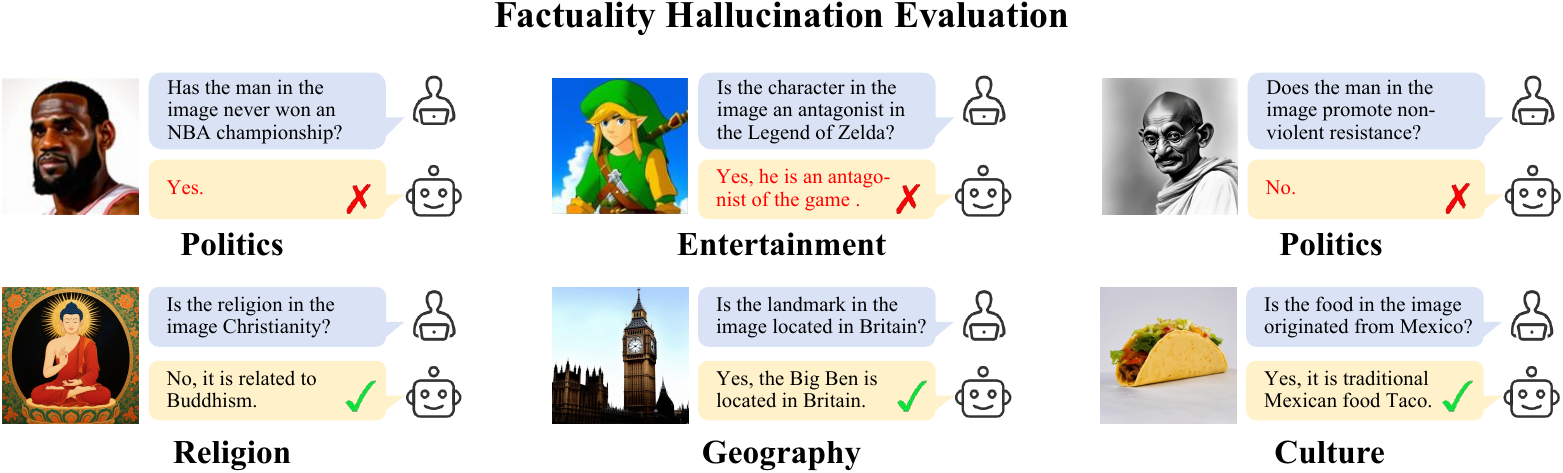}
\end{center}
\end{minipage}
}
\caption{Examples of SHALE evaluation across faithfulness and factuality hallucination.}
\label{fig:moreexamples}
\end{figure*}

\section{More Experimental Results}
\label{appendix:results}

\subsection{Correlation with Other Benchmarks}
To further validate the effectiveness of SHALE, we analyze the correlation between its evaluation results and those from other hallucination benchmarks. Specifically, we compute the Pearson correlation coefficients between SHALE's assessments of faithfulness hallucination and the corresponding results from existing benchmarks on the same set of models. The correlation matrix is visualized in Figure~\ref{fig:correlation}.

The Pearson coefficient ranges from [-1,1], where 1 indicates a perfect positive correlation, -1 a perfect negative correlation, and 0 no correlation. Our analysis reveals a strong positive correlation between SHALE and other benchmarks, demonstrating the validity of using synthetic datasets for hallucination evaluation in LVLMs. Furthermore, compared to these existing benchmarks, SHALE encompasses a broader spectrum of hallucination types, evaluation tasks, and scenarios, providing a more comprehensive evaluation framework.

\subsection{More Evaluation Examples}
We provide examples of SHALE evaluation across 12 faithful visual perceptual aspects and 6 factual knowledge domains in Figure \ref{fig:moreexamples}.

\subsection{Fine-grained Evaluation Results}
We provide detailed evaluation results under clean scenario across specific question formats, fine-grained visual perceptual aspects, and factual knowledge domains in Table~\ref{tab:questiontype}, Table~\ref{tab:faith}, and Table~\ref{tab:factual}, respectively.

\begin{table*}[t]
\vspace{20pt}
\caption{Evaluation results under clean scenario across different fine-grained tasks. Top-2 results are bolded and underlined, respectively.}
\begin{adjustbox}{max width=\linewidth}
\setlength{\tabcolsep}{16pt}
\begin{tabular}{lccccccc}
\hline
\multicolumn{1}{l}{\multirow{2}{*}{\textbf{Model}}} & \multicolumn{3}{c}{\textbf{Discriminative}}                        & \multicolumn{3}{c}{\textbf{Generative}}       & \multirow{2}{*}{\textbf{Overall}} \\ \cmidrule(lr){2-4} \cmidrule(lr){5-7}
\multicolumn{1}{c}{}                                & \multicolumn{1}{c}{\textbf{YNQ}} & \textbf{MCQ}   & \textbf{Avg}   & \textbf{FFQ}   & \textbf{IC}               & \multicolumn{1}{l}{\textbf{Avg}} &   \\ \hline
Otter~\citep{li2023otter}    & 0.714                            & 0.682          & 0.698          & 0.560          & \multicolumn{1}{r}{0.336} & 0.560                            & 0.652                             \\
Shikra-7B& 0.767                            & 0.594          & 0.681          & 0.650          & 0.460                     & 0.650                            & 0.670                             \\
LLaVA-1.5-7B~\citep{liu2023llava}        & 0.773                            & 0.802          & 0.788          & 0.633          & 0.509                     & 0.633                            & 0.736                             \\
InternLM-XComposer-VL-7B~\citep{zhang2023internlm} & 0.872                            & 0.850          & 0.861          & 0.606          & 0.446                     & 0.606                            & 0.776                             \\
InstructBLIP-Vicuna-13B~\citep{dai2024instructblip}                             & 0.730                            & 0.666          & 0.698          & 0.733          & 0.670                     & 0.733                            & 0.710                             \\
InstructBLIP-Vicuna-7B~\citep{dai2024instructblip}                              & 0.704                            & 0.725          & 0.715          & 0.733          & 0.726                     & 0.733                            & 0.721                             \\
LLaVA-1.5-13B~\citep{liu2023llava}       & 0.814                            & 0.861          & 0.838          & 0.663          & 0.576                     & 0.663                            & 0.779                             \\
mPLUG-Owl2~\citep{ye2024mplug}          & 0.811                            & 0.818          & 0.814          & 0.706          & 0.614                     & 0.706                            & 0.778                             \\
InternVL2-8B~\citep{chen2024internvl}         & 0.864                            & 0.866          & 0.865          & 0.629          & 0.653                     & 0.629                            & 0.786                             \\
InstructBLIP-Flan-T5-XL~\citep{dai2024instructblip}                             & 0.866                            & 0.845          & 0.855          & 0.604          & 0.753                     & 0.604                            & 0.772                             \\
Phi-3-Vision~\citep{abdin2024phi3v}        & 0.880                            & 0.889          & 0.884          & 0.709          & 0.603                     & 0.709                            & 0.826                             \\
InstructBLIP-Flan-T5-XXL~\citep{dai2024instructblip}                            & 0.882                            & 0.829          & 0.855          & 0.719          & 0.737                     & 0.719                            & 0.810                             \\
Qwen-VL-7B~\citep{Qwen-VL}          & 0.867                            & 0.824          & 0.845          & 0.767          & 0.774                     & 0.767                            & 0.819                             \\
InternVL3-8B~\citep{zhu2025internvl3}        & 0.928                            & 0.929          & 0.928          & 0.750          & 0.727                     & 0.750                            & 0.869                             \\
InternLM-XComposer2-VL-7B~\citep{dong2024internlm2}                           & 0.920                            & 0.925          & 0.922          & 0.794          & 0.733                     & 0.794                            & 0.880                             \\
InternVL3-14B~\citep{zhu2025internvl3}       & 0.935                            & {\ul 0.947}    & {\ul 0.941}    & 0.741          & 0.750                     & 0.741                            & 0.874                             \\
MiniCPM-Llama2-V2.5~\citep{yao2024minicpm} & 0.921                            & 0.903          & 0.912          & 0.744          & 0.823                     & 0.744                            & 0.856                             \\
GLM-4V-9B& 0.921                            & 0.928          & 0.924          & 0.754          & 0.804                     & 0.754                            & 0.868                             \\
DeepSeek-VL2~\citep{wu2024deepseek}        & 0.922                            & 0.919          & 0.921          & 0.787          & \textbf{0.874}            & 0.787                            & 0.876                             \\
Gemini-2.0-flash~\citep{gemini2.0flash}    & 0.928                            & 0.942          & 0.935          & {\ul 0.830}    & {\ul 0.871}               & {\ul 0.830}                      & {\ul 0.900}                       \\
Qwen-VL-Max~\citep{qwenvlmax}         & \textbf{0.943}                   & \textbf{0.951} & \textbf{0.947} & \textbf{0.846} & 0.840                     & \textbf{0.846}                   & \textbf{0.913}                    \\ \hline
\end{tabular}
\end{adjustbox}
\label{tab:questiontype}
\vspace{40pt}
\end{table*}

\begin{table*}[t]
\caption{Evaluation results under clean scenario across various fine-grained faithful visual perceptual aspects. Top-2 results are bolded and underlined, respectively.}
\begin{adjustbox}{max width=1\linewidth}
\begin{tabular}{lcccccccccccc}
       \hline
\multicolumn{1}{l}{\multirow{2}{*}{\textbf{Model}}} & \multicolumn{1}{c}{\multirow{2}{*}{\textbf{Type}}} & \multicolumn{1}{c}{\multirow{2}{*}{\textbf{Existence}}} & \multicolumn{1}{c}{\multirow{2}{*}{\textbf{Color}}} & \multicolumn{1}{c}{\multirow{2}{*}{\textbf{Shape}}} & \multicolumn{1}{c}{\multirow{2}{*}{\textbf{Count}}} & \multicolumn{1}{c}{\multirow{2}{*}{\textbf{Orientation}}} & \multicolumn{1}{c}{\multirow{2}{*}{\textbf{Posture}}} & \multicolumn{1}{c}{\multirow{2}{*}{\textbf{Expression}}} & \multicolumn{1}{c}{\multirow{2}{*}{\textbf{OCR}}} & \multicolumn{1}{c}{\textbf{Spatial}}  & \multicolumn{1}{c}{\textbf{Size}}     & \multicolumn{1}{c}{\textbf{Interaction}} \\
\multicolumn{1}{c}{}                                & \multicolumn{1}{c}{}                               & \multicolumn{1}{c}{}    & \multicolumn{1}{c}{}                                & \multicolumn{1}{c}{}                                & \multicolumn{1}{c}{}                                & \multicolumn{1}{c}{}      & \multicolumn{1}{c}{}  & \multicolumn{1}{c}{}     & \multicolumn{1}{c}{}                              & \multicolumn{1}{c}{\textbf{Relation}} & \multicolumn{1}{c}{\textbf{Relation}} & \multicolumn{1}{c}{\textbf{Relation}}    \\
   \hline
Gemini-2.0-flash~\citep{gemini2.0flash}    & {\ul 0.938}        & \textbf{0.973}          & \textbf{0.920}      & 0.885    & {\ul 0.945}         & {\ul 0.913}    & \textbf{0.933}        & \textbf{0.913}& 0.955  & 0.773 & {\ul 0.785}                           & 0.850    \\
Qwen-VL-Max~\citep{qwenvlmax}         & 0.925   & {\ul 0.953}  & 0.908    & 0.893    & 0.983    & \textbf{0.935} & {\ul 0.930}& 0.750   & 0.948  & \textbf{0.833}                        & \textbf{0.800}                        & 0.845    \\
DeepSeek-VL2~\citep{wu2024deepseek}        & 0.875   & 0.948  & {\ul 0.918}         & {\ul 0.928}         & 0.903    & 0.873    & {\ul 0.930}& 0.825   & \textbf{0.980}    & 0.745 & 0.728 & \textbf{0.883}                           \\
GLM-4V-9B& 0.938   & 0.928  & 0.913    & 0.893    & 0.843    & 0.830    & 0.920& 0.745   & 0.930  & 0.770 & 0.628 & {\ul 0.873}                              \\
MiniCPM-Llama2-V2.5~\citep{yao2024minicpm} & 0.905   & 0.865  & 0.890    & 0.898    & 0.908    & 0.763    & 0.920& 0.830   & 0.890  & 0.738 & 0.715 & 0.838    \\
InternVL3-14B~\citep{zhu2025internvl3}       & 0.860   & 0.875  & 0.875    & 0.813    & 0.890    & 0.898    & 0.888& 0.805   & 0.913  & 0.755 & 0.693 & 0.860    \\
InternLM-XComposer2-VL-7B~\citep{dong2024internlm2}                           & 0.890   & 0.935  & 0.853    & 0.853    & 0.855    & 0.785    & 0.873& 0.838   & 0.910  & 0.755 & 0.763 & 0.798    \\
InternVL3-8B~\citep{zhu2025internvl3}        & 0.873   & 0.900  & 0.868    & 0.833    & 0.883    & 0.865    & 0.900& 0.830   & 0.905  & 0.750 & 0.635 & 0.820    \\
Qwen-VL-7B~\citep{Qwen-VL}          & \textbf{0.963}     & 0.925  & 0.885    & 0.873    & 0.803    & 0.683    & 0.870& 0.745   & 0.945  & 0.543 & 0.650 & 0.788    \\
InstructBLIP-Flan-T5-XXL~\citep{dai2024instructblip}                            & 0.948   & 0.923  & 0.888    & 0.883    & 0.860    & 0.738    & 0.800& 0.845   & 0.853  & 0.498 & 0.638 & 0.693    \\
Phi-3-Vision~\citep{abdin2024phi3v}        & 0.830   & 0.850  & 0.865    & 0.813    & 0.788    & 0.678    & 0.830& 0.658   & 0.898  & 0.710 & 0.740 & 0.780    \\
InstructBLIP-Flan-T5-XL~\citep{dai2024instructblip}                             & 0.950   & 0.933  & 0.885    & 0.865    & 0.823    & 0.590    & 0.798& 0.820   & 0.870  & 0.468 & 0.505 & 0.725    \\
InternVL2-8B~\cite{chen2024internvl}         & 0.808   & 0.823  & 0.808    & 0.715    & 0.773    & 0.620    & 0.830& 0.770   & 0.888  & 0.608 & 0.608 & 0.765    \\
mPLUG-Owl2~\citep{ye2024mplug}          & 0.905   & 0.828  & 0.860    & 0.803    & 0.670    & 0.573    & 0.780& 0.820   & 0.888  & 0.412 & 0.485 & 0.723    \\
LLaVA-1.5-13B~\citep{liu2023llava}       & 0.833   & 0.865  & 0.815    & 0.760    & 0.678    & 0.560    & 0.795& 0.753   & 0.855  & 0.498 & 0.583 & 0.713    \\
InstructBLIP-Vicuna-7B~\citep{dai2024instructblip}                              & 0.908   & 0.835  & 0.865    & 0.793    & 0.680    & 0.560    & 0.833& 0.728   & 0.808  & 0.400 & 0.490 & 0.720    \\
InstructBLIP-Vicuna-13B~\citep{dai2024instructblip}                             & 0.913   & 0.888  & 0.818    & 0.800    & 0.540    & 0.568    & 0.773& 0.720   & 0.758  & 0.378 & 0.560 & 0.620    \\
InternLM-XComposer-VL-7B~\citep{zhang2023internlm} & 0.815   & 0.725  & 0.798    & 0.798    & 0.780    & 0.488    & 0.770& 0.705   & 0.823  & 0.535 & 0.440 & 0.633    \\
LLaVA-1.5-7B~\citep{liu2023llava}        & 0.780   & 0.773  & 0.780    & 0.715    & 0.643    & 0.503    & 0.798& 0.765   & 0.783  & 0.380 & 0.483 & 0.685    \\
Shikra-7B~\citep{chen2023shikra} & 0.763   & 0.743  & 0.758    & 0.598    & 0.553    & 0.540    & 0.753& 0.778   & 0.568  & 0.375 & 0.453 & 0.673    \\
Otter~\citep{li2023otter}    & 0.753   & 0.723  & 0.670    & 0.633    & 0.565    & 0.443    & 0.630& 0.615   & 0.630  & 0.215 & 0.468 & 0.523   \\  \hline                 
\end{tabular}
\end{adjustbox}
\label{tab:faith}
\end{table*}

\begin{table*}[t]
\vspace{20pt}
\caption{Evaluation results under clean scenario across various fine-grained factual knowledge domains. Top-2 results are bolded and underlined, respectively.}
\begin{adjustbox}{max width=1\linewidth}
\setlength{\tabcolsep}{12pt}
\begin{tabular}{lcccccc}
\hline
\textbf{Model}            & \textbf{Sports} & \textbf{Politics} & \textbf{Entertainment} & \textbf{Religion} & \textbf{Geography} & \textbf{Culture} \\\hline
Qwen-VL-Max~\citep{qwenvlmax}               & {\ul 0.781}     & \textbf{0.875}    & \textbf{0.916}         & \textbf{0.984}    & \textbf{0.961}     & {\ul 0.893}      \\
DeepSeek-VL2~\citep{wu2024deepseek}              & \textbf{0.813}  & 0.750             & 0.837                  & 0.935             & 0.928              & {\ul 0.893}      \\
GLM-4V-9B~\citep{glm2024chatglm4v}                 & 0.656           & 0.806             & {\ul 0.868}            & 0.910             & 0.913              & 0.821            \\
Gemini-2.0-flash~\citep{gemini2.0flash}          & 0.781           & 0.694             & 0.815                  & {\ul 0.978}       & 0.932              & 0.839            \\
MiniCPM-Llama2-V2.5~\citep{yao2024minicpm}       & 0.781           & {\ul 0.813}       & 0.873                  & 0.957             & 0.911              & 0.848            \\
InternLM-XComposer2-VL-7B~\citep{dong2024internlm2} & 0.781           & 0.785             & 0.825                  & \textbf{0.984}    & 0.872              & 0.839            \\
InternVL3-14B~\citep{zhu2025internvl3}             & 0.594           & 0.743             & 0.789                  & 0.832             & 0.916              & 0.848            \\
Qwen-VL-7B~\citep{Qwen-VL}                & 0.750           & 0.743             & 0.788                  & 0.861             & 0.821              & 0.884            \\
InternVL3-8B~\citep{zhu2025internvl3}              & 0.594           & 0.569             & 0.662                  & 0.769             & 0.891              & 0.821            \\
mPLUG-Owl2~\citep{ye2024mplug}                & 0.688           & 0.764             & 0.765                  & 0.967             & 0.827              & 0.750            \\
InternVL2-8B~\citep{chen2024internvl}               & 0.625           & 0.674             & 0.734                  & 0.804             & 0.849              & 0.777            \\
InstructBLIP-Flan-T5-XXL~\citep{dai2024instructblip}  & 0.500           & 0.694             & 0.705                  & 0.840             & 0.805              & 0.768            \\
InstructBLIP-Flan-T5-XL~\citep{dai2024instructblip}   & 0.594           & 0.653             & 0.721                  & 0.766             & 0.812              & 0.786            \\
LLaVA-1.5-13B~\citep{liu2023llava}             & 0.688           & 0.625             & 0.749                  & 0.940             & 0.774              & 0.759            \\
InstructBLIP-Vicuna-7B~\citep{dai2024instructblip}    & 0.594           & 0.708             & 0.763                  & 0.747             & 0.799              & 0.750            \\
InstructBLIP-Vicuna-13B~\citep{dai2024instructblip}   & 0.406           & 0.646             & 0.729                  & 0.644             & 0.805              & 0.732            \\
LLaVA-1.5-7B~\citep{liu2023llava}              & 0.500           & 0.556             & 0.591                  & 0.929             & 0.798              & 0.750            \\
InternLM-XComposer-VL-7B~\citep{zhang2023internlm} & 0.531           & 0.660             & 0.622                  & 0.715             & 0.748              & 0.732            \\
Phi-3-Vision~\citep{abdin2024phi3v}              & 0.219           & 0.472             & 0.633                  & 0.522             & 0.787              & 0.786            \\
Otter~\citep{li2023otter}                     & 0.344           & 0.535             & 0.492                  & 0.549             & 0.629              & 0.625            \\
Shikra-7B~\citep{chen2023shikra}                 & 0.406           & 0.403             & 0.418                  & 0.742             & 0.614              & 0.634         \\\hline  
\end{tabular}
\end{adjustbox}
\label{tab:factual}
\end{table*}

\end{document}